\documentclass{article}

\usepackage{PRIMEarxiv}

\usepackage[utf8]{inputenc}          
\usepackage[T1]{fontenc}             
\usepackage[hidelinks]{hyperref}     
\usepackage{url}
\usepackage{booktabs}
\usepackage{amsmath}
\usepackage{amsthm}
\usepackage{amsfonts}
\usepackage{times}
\usepackage{soul}
\usepackage{graphicx}
\usepackage{algorithm}
\usepackage{algorithmic}
\usepackage{color}
\usepackage{dsfont}
\usepackage{multirow}
\usepackage{stfloats}
\usepackage[flushleft]{threeparttable}

\newtheorem{theorem}{Theorem}

\makeatletter
\newtheorem*{rep@theorem}{\rep@title}
\newcommand{\newreptheorem}[2]{%
\newenvironment{rep#1}[1]{%
 \def\rep@title{#2 \ref{##1}}%
 \begin{rep@theorem}}%
 {\end{rep@theorem}}}
\makeatother

\newreptheorem{theorem}{Theorem}

\title{Graph is all you need? Lightweight data-agnostic neural architecture search without training}

\author{
  Zhenhan Huang \\
  Rensselaer Polytechnic Institute\\
  \And
  Tejaswini Pedapati \\
  IBM Research\\
  \And
  Pin-Yu Chen \\
  IBM Research\\
  \And
  Chunheng Jiang \\
  Rensselaer Polytechnic Institute\\
  \And
  Jianxi Gao\thanks{Corresponding author: gaoj8@rpi.edu} \\
  Rensselaer Polytechnic Institute\\
}

\graphicspath{{figs/}}                 

\definecolor{RoyalBlue}{rgb}{0.25, 0.41, 0.88}
\definecolor{DarkBlue}{rgb}{0.0, 0.0, 0.55}
\definecolor{DarkGray}{rgb}{0.66, 0.66, 0.66}

\def\nasgraph{\textit{NASGraph}}

\begin{document}
\maketitle

\begin{abstract}

Neural architecture search (NAS) enables the automatic design of neural network models. However, training the candidates generated by the search algorithm for performance evaluation incurs considerable computational overhead. Our method, dubbed \nasgraph{}, remarkably reduces the computational costs by converting neural architectures to graphs and using the average degree, a graph measure, as the proxy in lieu of the evaluation metric. Our training-free NAS method is data-agnostic and light-weight. It can find the best architecture among 200 randomly sampled architectures from NAS-Bench201 in \textbf{217 CPU seconds}. Besides, our method is able to achieve competitive performance on various datasets including NASBench-101, NASBench-201, and NDS search spaces. We also demonstrate that \nasgraph{} generalizes to more challenging tasks on Micro TransNAS-Bench-101. 

\end{abstract}

\section{Introduction}

Neural architecture search (NAS) aims to automate the process of discovering state-of-the-art (SOTA) deep learning models. The objective of NAS is to find an optimal neural architecture $a^* = \arg\min_{a \in \mathcal{A}}f(a)$, where $f(a)$ denotes the performance (e.g., a task-specific loss function) of the neural architecture $a$ trained for a fixed number of epochs using a dataset, and $\mathcal{A}$ is the search space. Previous NAS techniques achieve competitive performance in various applications such as image classification \cite{Zoph2018}, object detection \cite{chen2019detnas} and semantic segmentation \cite{liu2019auto}. The pioneering work \cite{zoph2016neural} based on reinforcement learning is resource intensive, requiring more than 20k GPU days. In particular, training the generated neural architecture candidates to evaluate their performance is the most computationally expensive step. To accelerate the search process, various proposals have been made, including weight sharing \cite{Pham2018,chu2021fairnas}, progressive complexity search stage \cite{Liu2018}, gradient descent in the differentiable search space \cite{liu2018darts}, predictor-based NAS techniques \cite{white2021bananas},  etc. The recent emergence of training-free NAS \cite{mellor2021neural,abdelfattah2021zero} pushes the boundary of efficient NAS techniques further and greatly eases the computational burden. Training-Free NAS computes a proxy metric in place of accuracy to rank the candidate architectures. The proxy metric is obtained by a single forward/backward propagation using a training dataset.

In this paper, we take a novel perspective on mapping neural networks to graphs: we treat inputs to neural components as graph nodes and use relationship between inputs to neighboring graph components to determine the connectivity between pairs of graph nodes. In this way, we are able to convert a neural architecture to a DAG $G(V, E)$ where node set $V = \{v_1, v_2, \ldots, v_n\}$ and edge set $E = \{e_{ij}\} \; \forall i, j$ s.t. there exists an edge from $v_{i}$ to $v_{j}$. After establishing the mapping, we extract the associated graph measures as NAS metrics to rank neural architectures. We note that the entire process is training-free and data-agnostic (i.e. do not require any training data).

We summarize our \textbf{main contributions} as follows:
\begin{itemize}
  \item We propose \nasgraph{}, a training-free and data-agnostic method for NAS. \nasgraph{} maps the neural architecture space to the graph space. To our best knowledge, this is the \textbf{first work} to apply graph theory for NAS. 
  \item Using the extracted graph measures for NAS, \nasgraph{} achieves competitive performance on NAS-Bench-101, NAS-Bench-201, Micro TransNAS-Bench-101 and NDS benchmarks, when compared to existing training-free NAS methods. The analysis of bias towards operations in the benchmarks indicates that \nasgraph{} has the lowest bias compared to those methods.
  \item In comparison to existing training-free NAS techniques, we show that the computation of \nasgraph{} is lightweight (only requires CPU).
\end{itemize}

\begin{figure*}[ht]
  \centering
  \includegraphics[width=.74\linewidth]{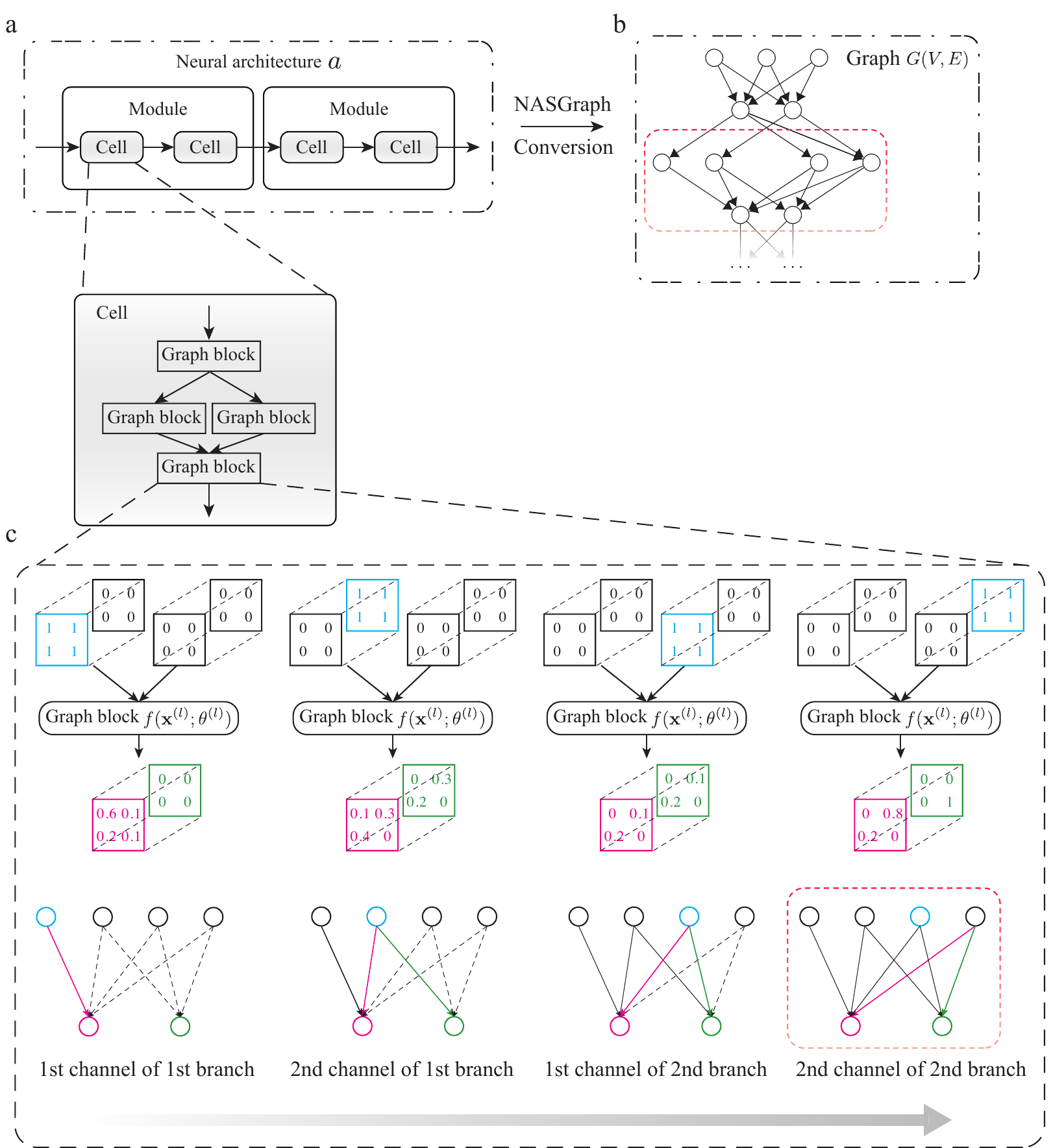}
  \caption{An overview of the \nasgraph{} framework: the connectivity of graph nodes is determined by the forward propagation of the corresponding graph blocks. In the toy example shown in the bottom of the figure, if the output from the forward propagation is all-zeros matrix $\mathds{O}$, there is no connection. Otherwise, the connection is built between a pair of graph nodes. The orange rectangles in (b) and (c) mark how a subgraph generated by a single forward propagation constitutes a part of the whole graph.}
  \label{fig:supernode_agg}
\end{figure*}

\section{Related Work}

\textbf{One-shot NAS.} One-shot NAS constructs a supernet subsuming all candidate architectures in the search space. In other words, subnetworks of the supernet are candidate architectures. The supernet method is faster than the conventional NAS methods because it enables weight sharing among all the candidate architectures in the search space. The first one-shot NAS algorithm \cite{Pham2018} considers the operations at each edge to be a discrete decision. DARTS \cite{liu2018darts} devises a continuous relaxation by formulating the value of the operation applied to an edge as a weighted sum of all the operations. Several works, e.g.\cite{xu2019pc}, \cite{wang2021_dartspt}, \cite{zela2020_robust_darts} were developed to improve some shortcomings of DARTS such as memory consumption, DARTS collapse and increase in the validation loss after the discretization step.
\cite{yu2020_Evaluating} demonstrated the discrepancy between the ranking of architectures sampled from a trained supernet and that of the same architectures when trained from scratch. To alleviate co-adaptation among the operations, works such as \cite{Guo2020_spos},  \cite{Bender2018_understanding}, recommend uniformly sampling the candidate architectures while others recommended pruning the search space during the search \cite{Noy2020_asap}. To accelerate the search process, EcoNAS \cite{zhou2020econas} proposes a hierarchical proxy strategy that uses a faster proxy for less promising candidates and a better but slower proxy for more promising candidates.


\textbf{Training-Free NAS.} Training-Free NAS uses models with randomly initialized weights to obtain the saliency metrics that rank these models. Since there is no need for training models, this routine is considerably faster even compared to one-shot NAS. NASWOT \cite{mellor2021neural} applies the theory on the linear regions in deep networks \cite{hanin2019complexity} to achieve NAS without training. The saliency metrics for the pruning-at-initialization work in network pruning are also found to be effective in zero-cost NAS \cite{abdelfattah2021zero}. In addition, TENAS \cite{chen2021neural} uses metrics from the condition number of neural tangent kernel and the number of linear regions. A comparison of saliency metrics \cite{krishnakumar2022bench} finds that these metrics might contain complementary information and hence a combination of metrics can be helpful in NAS. 

\section{NASGraph: A Graph-Based Framework for Data-Agnostic and Training-Free NAS}

Figure \ref{fig:supernode_agg} shows  an overview of our proposed \nasgraph{} framework. A neural architecture is uniquely mapped to a DAG, i.e. $a \mapsto G(V, E)$. After the conversion, graph measures are computed to rank the performance of the corresponding neural architectures. We also note that our notion of graph refers to the specific graph representation of a neural architecture $a$ via our proposed graph conversion techniques, instead of the original computation graph of $a$.


\textbf{Graph block.} The basic element in the \nasgraph{} framework is the graph block. We use the notation $f^{(l)}(\mathbf{x}^{(l)}; \mathbf{\theta}^{(l)})$ to represent the $l$-th graph block, where $\mathbf{x}^{(l)}$ is the input to the graph block and $\mathbf{\theta}^{(l)}$ is the model parameter. We combine several neural components (e.g. Conv and ReLU) to a graph block such that the output of the graph block $y^{(l)} = f^{(l)}(\mathbf{x}^{(l)}; \mathbf{\theta}^{(l)})$ is non-negative.



\textbf{Conversion method.} Inspired by the iterative re-evaluation method in the network pruning \cite{verdenius2020pruning,tanaka2020pruning}, we apply conversion to each graph block independently. Similarly, we also use all-ones matrix as the input to the graph block $\mathbf{x}^{(l)} = \mathds{1}^{C^{(l-1)}\times H^{(l-1)} \times W^{(l-1)}}$ in the forward propagation process, where $C^{(l-1)}$ is the number of channels, $H^{(l-1)}$ is the image height, and $W^{(l-1)}$ is the image width for the input to $l$-th graph block. This helps us get an unbiased estimate of the contribution of the input to the output. Further, to determine the contribution of the $c$-th channel of the input on the channels of the output for the $l$-th graph block, we apply a mask $\mathcal{M}^{(l)}_{c}$ to the input so that only the $c$-th channel $(\mathbf{x}^{(l)}_{d_1d_2d_3})_{d_1=c}$ is an all-ones matrix $\mathds{1}^{H^{(l-1)} \times W^{(l-1)}}$ and other channels are zero matrices $\mathds{O}^{H^{(l-1)} \times W^{(l-1)}}$. A toy example is shown in the bottom of Figure \ref{fig:supernode_agg}. We evaluate the contribution of the $c$-th channel $(\mathbf{x}^{(l)}_{d_1d_2d_3})_{d_1=c}$ to the output $\mathbf{y}^{(l)}$ by performing a forward propagation as described by:
\begin{equation}
  \mathbf{y}_{c}^{(l)} = f^{(l)}(\mathcal{M}^{(l)}_{c} \odot \mathbf{x}^{(l)}; \mathbf{\theta}^{(l)})
\end{equation}
where $f^{(l)}(\cdot)$ is the $l$-th graph block, $\odot$ is the Hadamard product, and $\mathbf{\theta}^{(l)}$ is the parameters of the $l$-th graph block. The score $\omega_{i^{(l-1)}j^{(l)}}$ for the edge $e_{ij}$ between node $i^{(l-1)}$ and $j^{(l)}$ is determined by:
\begin{equation}\label{eq:agg}
  \omega_{i^{(l-1)}j^{(l)}} = \sum_{d_{2}=1}^{H^{(l)}}\sum_{d_{3}=1}^{W^{(l)}} (({y_{i}^{(l)}})_{d_1d_2d_3})_{d_1 = j}
\end{equation}
If $\omega_{i^{(l-1)}j^{(l)}}$ is larger than 0, we build an edge between node $i^{(l-1)}$ and node $j^{(l)}$ that indicates the relationship between $i$-th channel of the input $\mathbf{x}^{(l)}$ and $j$-th channel of the output $\mathbf{y}^{(l)}$.
We use a virtual input graph block of identity operation to take the input to the neural architecture into consideration. After looping though all graph blocks, we can uniquely construct a graph.

\textbf{Analysis of the proposed method.} We consider a neural component consisting of convolution and ReLU activation function. We use $\mathcal{F}$ to represent the neural component, $\mathcal{F}(\mathbf{x}) = h \circ f(\mathbf{x})$. The function $f(\cdot)$ is the convolution operation and $h(\cdot)$ is the ReLU activation function.

\begin{theorem}
    \label{th:conv_relu}
    Let $\mathbf{x}$ be the input to $\mathcal{F}$. $\mathcal{F}$ is converted to a graph $G(V, E)$ using NASGraph framework: the input to $\mathcal{F}$ has $\mathds{1} \in \mathbb{R}^{H^{(0)} \times W^{(0)}}$ for the channel $i$ and $\mathds{O} \in \mathbb{R}^{H^{(0)} \times W^{(0)}}$ for rest of channels. There is no edge between node i and node j if and only if the output for the channel j is all-zeros matrix $\mathds{O} \in \mathbb{R}^{H^{(1)} \times W^{(1)}}$.
\end{theorem}

We provide the proof in Appendix \ref{append:conv_relu}. Recall that we use $\mathds{1}$ as input to $\mathcal{F}$ to make an unbiased estimate on the contribution of the input to the output. Hence, an output of $\mathds{O}$ means there exists a contribution from the input to the output, which is inferior to the performance of neural architecture. Theorem \ref{th:conv_relu} indicates that the connectivity between graph nodes is associated with the neural architecture performance. Hence, we can use graph measures that are sensitive to node connectivity as the metrics to rank neural architectures.

\begin{table}[t]
  \centering
  \caption{Comparison of the ranking correlation between \nasgraph{} and training-free NAS methods using single metric on NAS-Bench-201. Correlations are calculated between the metrics and test accuracies.}\label{table:nasbench201_comp}
  \resizebox{.7\textwidth}{!} {%
  \begin{tabular}{ccccccccccc}
    \hline
    \multirow{2}*{Method} & \multirow{2}*{Metric} & & \multicolumn{2}{c}{CIFAR-10} & & \multicolumn{2}{c}{CIFAR-100} & & \multicolumn{2}{c}{ImageNet-16-120} \\
    \cline{4-5}\cline{7-8}\cline{10-11}
     & & & $\rho$ & $\tau$ & & $\rho$ & $\tau$ & & $\rho$ & $\tau$ \\
     \hline
    NASWOT & \texttt{naswot} & & 0.76 & 0.57 & & 0.79 & \textbf{0.61} & & 0.71 & 0.55 \\
    \hline
    ZiCo$^{\ddagger}$ & \texttt{zico} & & 0.74 & 0.55 & & 0.78 & 0.58 & & 0.76 & 0.56 \\
    \hline
    \multirow{2}*{TENAS} & NTK & & - & - & & - & -0.42 & & - & - \\
     & NLR & - & - & & - & -0.50 & & - & - & -\\
    \hline
    \multirow{6}*{Zero-Cost NAS} & \texttt{grad\_norm} & & 0.58 & 0.42 & & 0.64 & 0.47 & & 0.58 & 0.43 \\
     & \texttt{snip} & & 0.58 & 0.43 & & 0.63 & 0.47 & & 0.58 & 0.43 \\
     & \texttt{grasp} & & 0.48 & 0.33 & & 0.54 & 0.38 & & 0.56 & 0.40 \\
     & \texttt{fisher} & & 0.36 & 0.26 & & 0.39 & 0.28 & & 0.33 & 0.25 \\
     & \texttt{synflow} & & 0.74 & 0.54 & & 0.76 & 0.57 & & 0.75 & 0.56 \\
     & \texttt{jacob\_cov} & & 0.73 & 0.55 & & 0.71 & 0.55 & & 0.71 & 0.54 \\
    \hline
    Ours & \texttt{avg\_deg} & & \textbf{0.78} & \textbf{0.58} & & \textbf{0.80} & 0.60 & & \textbf{0.77} & \textbf{0.57} \\
    \hline
  \end{tabular}%
  }
  
  \begin{tablenotes}
    \footnotesize
    \item $\ddagger$ Original implementation of ZiCo uses cutout data augmentation. To make a fair comparison, we recalculate the correlation without cutout data augmentation.
  \end{tablenotes}
\end{table}

\textbf{Improving the search efficiency of NASGraph.}
To reduce computational overhead, NAS typically uses a training-reduced proxy to obtain the performance of neural architectures. A systematic study is reported in EcoNAS \cite{zhou2020econas} where four reducing factors are analyzed: (1) number of epochs, (2) resolution of input images, (3) number of training samples, (4) number of channels for Convolution Neural Networks (CNNs). Following their convention, to accelerate \nasgraph{}, we also consider the surrogate models, i.e. models with computationally reduced settings. We dub the surrogate model \textbf{\nasgraph{(h, c, m)}}, where $h$ is the number of channels, $c$ is the number of search cells in a module, and $m$ is the number of modules. The ablation study on the effect of using a surrogate model is discussed in Appendix \ref{append:ablation}.

After converting the neural architectures to graphs, we use the average degree as the graph measure to rank neural architectures. In addition to the average degree, we also examine the performance of other graph measures. The result is included in the Appendix \ref{append:graph_measures}. 

\section{Performance Evaluation}

\begin{table}[t]
  \centering
  \caption{Comparison of Kendall's Tau ranking correlations $\tau$ between test accuracies and training-free NAS metrics on the NDS benchmark.}\label{table:single_metric_nds}
  \resizebox{0.6\textwidth}{!}{%
  \begin{tabular}{cccccc}
    \hline
     Metric & AMOEBA & DARTS & ENAS & NASNet & PNAS \\
    \hline
    \texttt{grad\_norm} & -0.12 & 0.22 & 0.06 & -0.05 & 0.15 \\
    \texttt{snip} & -0.09 & 0.26 & 0.10 & -0.02 & 0.18 \\
    \texttt{grasp} & 0.02 & -0.04 & 0.03 & 0.18 & -0.01 \\
    \texttt{fisher} & -0.12 & 0.19 & 0.04 & -0.07 & 0.14 \\
    \texttt{jacov\_cov} & 0.22 & 0.19 & 0.11 & 0.05 & 0.10 \\
    \texttt{synflow} & -0.06 & 0.30 & 0.14 & 0.04 & 0.21 \\
    \texttt{naswot} & 0.22 & \textbf{0.47} & 0.37 & 0.30 & 0.38 \\
    \hline
    \texttt{avg\_deg} (Ours) & \textbf{0.32} & 0.45 & \textbf{0.41} & \textbf{0.37} & \textbf{0.40} \\
    \hline
  \end{tabular}
  }
\end{table}

\begin{table}[t]
  \centering
  \caption{Comparison of Spearman's ranking correlations $\rho$ between validation accuracies and training-free NAS metrics on micro TransNAS-Bench-101. The baseline performance is extracted from \protect\cite{krishnakumar2022bench}. Note that \texttt{synflow} and \texttt{avg\_deg} are data-agnostic. CO is \texttt{class\_object}, CS is \texttt{class\_scene}, RL is \texttt{room\_layout}, SS is \texttt{segment\_semantic}.}\label{table:single_metric_transbench101}
  \resizebox{.4\textwidth}{!}{%
  \begin{tabular}{cccccc}
    \hline
    \multirow{2}*{Metric} & \multicolumn{4}{c}{Micro TransNAS-Bench-101} \\
    \cline{2-5}
     & CO & CS & RL & SS \\
    \hline
    \texttt{plain} & 0.34 & 0.24 & 0.36 & -0.02 \\
    \texttt{grasp} & -0.22 & -0.27 & -0.29 & 0.00 \\
    \texttt{fisher} & 0.44 & 0.66 & 0.30 & 0.12 \\
    \texttt{epe\_nas} & 0.39 & 0.51 & \textbf{0.40} & 0.00 \\
    \texttt{grad\_norm} & 0.39 & 0.65 & 0.25 & 0.60 \\
    \texttt{snip} & 0.45 & 0.70 & 0.32 & 0.68 \\
    \texttt{synflow} & 0.48 & 0.72 & 0.30 & 0.00 \\
    \texttt{l2\_norm} & 0.32 & 0.53 & 0.18 & 0.48 \\
    \texttt{params} & 0.45 & 0.64 & 0.30 & 0.68 \\
    \texttt{zen} & 0.54 & 0.72 & 0.38 & 0.67 \\
    \texttt{jacob\_cov} & 0.51 & \textbf{0.75} & \textbf{0.40} & \textbf{0.80} \\
    \texttt{flops} & 0.46 & 0.65 & 0.30 & 0.69 \\
    \texttt{naswot} & 0.39 & 0.60 & 0.25 & 0.53 \\
    \texttt{zico} & 0.55 & 0.68 & 0.26 & 0.61 \\
    \hline
    \texttt{avg\_deg} (Ours) & \textbf{0.55} & 0.70 & 0.37 & 0.66 \\
    \hline
  \end{tabular}
  }
\end{table}

\subsection{Experiment Setup}

\textbf{NAS benchmarks.} To examine the effectiveness of our proposed \nasgraph{}, we use neural architectures on NAS-Bench-101 \cite{ying2019bench}, NAS-Bench-201 \cite{dong2020bench}, TransNAS-Bench-101 \cite{duan2021transnas} and Network Design Space (NDS) \cite{radosavovic2019network}. NAS-Bench-101 is the first NAS benchmark consisting of 423,624 neural architectures trained on the CIFAR-10 dataset \cite{Krizhevsky2009}. The training statistics are reported at 108th epoch. NAS-Bench-201 is built for prototyping NAS algorithms. It contains 15,625 neural architectures trained on CIFAR-10, CIFAR-100 \cite{Krizhevsky2009} and ImageNet-16-120 \cite{chrabaszcz2017downsampled} datasets. The training statistics are reported at 200th epoch for these three datasets. In addition to standard NAS benchmarks, we also examine the performance of \nasgraph{} on TransNAS-Bench-101, specifically the micro search space. The micro (cell-level) TransNAS-Bench-101 has 4,096 architectures trained on different tasks using the Taskonomy dataset \cite{zamir2018taskonomy}. The NDS benchmark includes AmoebaNet \cite{real2019regularized}, DARTS \cite{liu2018darts}, ENAS \cite{pham2018efficient}, NASNet \cite{zoph2016neural} and PNAS search spaces \cite{liu2018progressive}.

We convert neural architectures with Gaussian initialization to graphs $G(V, E)$.  The conversion is repeated 8 times with different random initializations. We find the difference in the rankings among 8 initializations is considerably marginal as shown in Appendix \ref{append:rank_var}.

We use AMD EPYC 7232P CPU in the computation of \nasgraph{}. To compute the performance of baselines requiring GPUs, a single NVIDIA A40 GPU is used. We reduce the number of channels to be 16 and the number of cells to be 1, i.e. \nasgraph{(16, 1, 3)} is used as the surrogate model.

\textbf{Baselines.} We use metrics in training-free NAS as our baselines. \texttt{zico} \cite{li2023zico} is based on the theory of Gram Matrix \cite{du2018gradient}. \texttt{relu\_logdet} (also dubbed \texttt{naswot}) \cite{mellor2021neural} applies the theory on the number of linear regions to represent the model expressivity. \texttt{jacob\_cov} \cite{mellor2021neural} is based on the correlation of Jacobians with inputs. \texttt{grad\_norm} \cite{abdelfattah2021zero} sums the Euclidean norm of the gradients. \texttt{snip} \cite{lee2018snip} is related to the connection sensitivity of neural network model. \texttt{grasp} \cite{wang2020picking} is based on the assumption that gradient flow is preserved in the efficient training.  \texttt{fisher} \cite{theis2018faster} estimates fisher information of model parameters, \texttt{synflow} \cite{tanaka2020pruning} preserves the total flow of synaptic strength.


\begin{table}[t]
  \centering
  \caption{Comparison of the ranking correlations using multiple metrics with training-free NAS methods on NAS-Bench-201. Correlations between the combined metric and NAS Benchmark performance are reported. TENAS \protect\cite{chen2021neural} combines rankings by NTK and NLR. Zero-Cost NAS \protect\cite{abdelfattah2021zero} takes a majority vote among the three metrics: \texttt{synflow}, \texttt{jacob\_cov} and \texttt{snip}. Our method combines the rankings of \texttt{avg\_deg} and \texttt{jacob\_cov}.}\label{table:multiple_metric_compare}
  \resizebox{.7\linewidth}{!} {%
  \begin{tabular}{cccccccccc}
    \hline
    \multicolumn{2}{c}{\multirow{2}*{Method}} & \multicolumn{2}{c}{CIFAR-10} & & \multicolumn{2}{c}{CIFAR-100} & & \multicolumn{2}{c}{ImageNet-16-120} \\
    \cline{3-4}\cline{6-7}\cline{9-10}
     & & $\rho$ & $\tau$ & & $\rho$ & $\tau$ & & $\rho$ & $\tau$ \\
    \hline
    TENAS & Rank combine & - & - & & - & 0.64 & & - & -\\
    Zero-Cost NAS & Voting & 0.82 & - & & 0.83 & - & & 0.82 & - \\
    Ours & Rank combine & \textbf{0.85} & 0.66 & & \textbf{0.85} & \textbf{0.67} & & \textbf{0.82} & 0.64\\
    \hline
  \end{tabular}
  }
\end{table}

\begin{table}[t]
    \centering
    \caption{Comparison of the accumulated frequency difference between training-free NAS methods and GT on top 10\% architectures of the NAS-Bench-201. GT ranks architectures ranked by test accuracies. Lower value means less bias (i.e. closer to GT).}\label{table:freq_pref_compare}
  \resizebox{0.7\textwidth}{!}{%
  \begin{tabular}{cccccccccc}
    \hline
    Metric & CIFAR-10 & CIFAR-100 & ImageNet-16-120 & Average bias \\
    \hline
    \texttt{relu\_logdet} & 0.3  & 0.27 & 0.19 & 0.25 \\
    \texttt{grad\_norm}   & 0.32 & 0.3  & 0.24 & 0.29 \\
    \texttt{snip}         & 0.31 & 0.29 & 0.24 & 0.28 \\
    \texttt{grasp}        & 0.31 & 0.28 & 0.24 & 0.28 \\
    \texttt{fisher}       & 0.52 & 0.52 & 0.53 & 0.52 \\
    \texttt{synflow}      & 0.22 & 0.18 & 0.27 & 0.22 \\
    \texttt{jacob\_cov}   & 0.39 & 0.42 & 0.22 & 0.34 \\
    \hline
    \multicolumn{5}{c}{Our method} \\
    \texttt{avg\_deg}     & 0.22 & \textbf{0.14} & 0.27 & 0.21 \\
    \texttt{comb\_rank}   & \textbf{0.17} & 0.17 & \textbf{0.12} & \textbf{0.15} \\
    \hline
  \end{tabular}
  }
\end{table}

\subsection{Scoring neural architectures using average degree}

\begin{figure*}[ht]
  \centering
  \includegraphics[width=\linewidth]{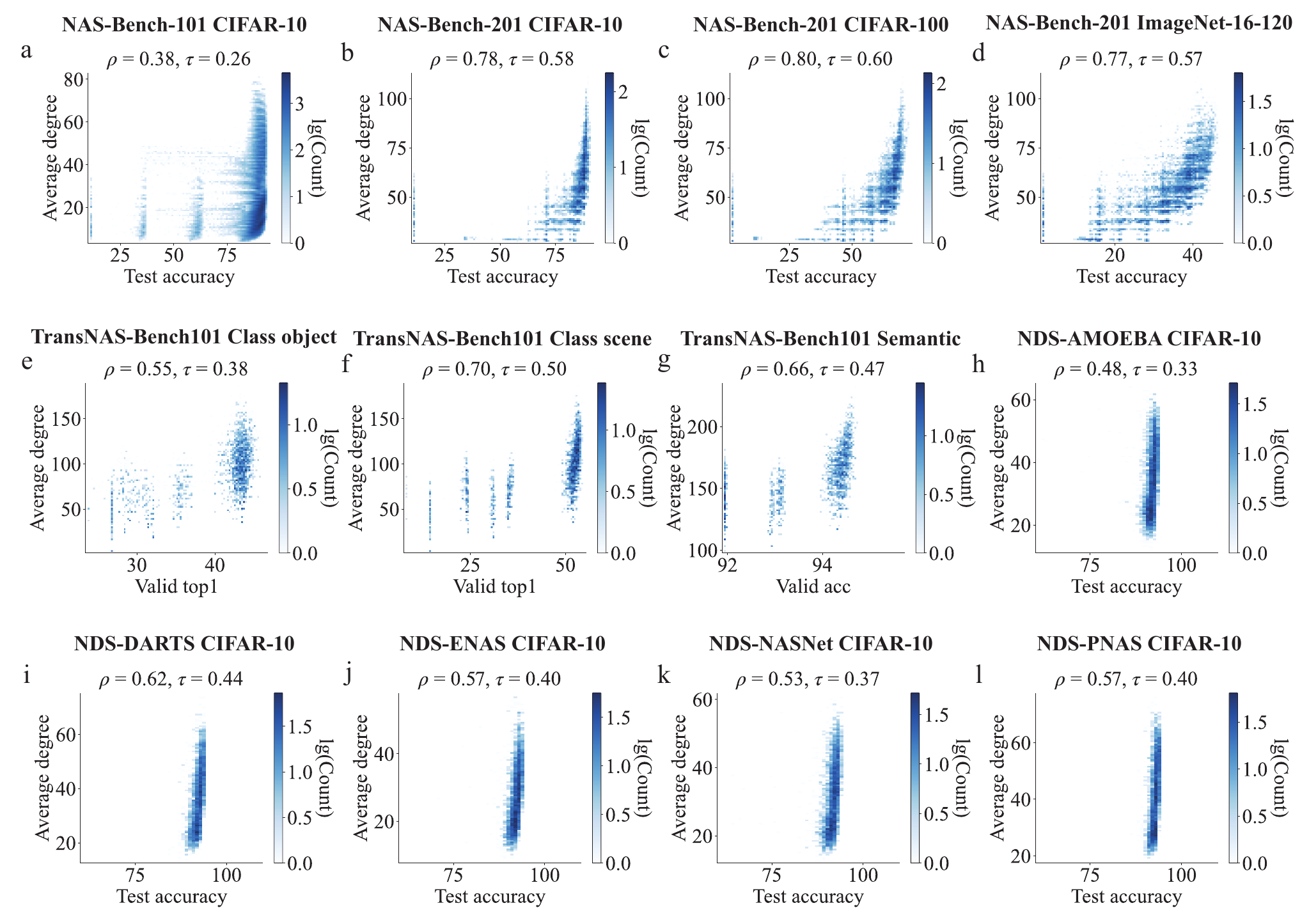}
  \vspace{-1em}
  \caption{The ranking correlation between the performance of neural architecture $a$ and the graph measures of the corresponding graph $G(V, E)$. In the \nasgraph{} framework, each neural architecture is uniquely mapped to a graph, i.e. $a \mapsto G(V, E)$.}
  \label{fig:nasbenchmark_corr}
\end{figure*}

The distribution of the average degree of graphs vs the performance of neural architectures is plotted in Figure \ref{fig:nasbenchmark_corr} (a)-(l). Colorbar indicates the number of architectures in logarithmic base 10. We compute both Spearman's ranking correlation $\rho$ and Kendall's Tau ranking correlation $\tau$ between them. In Table \ref{table:nasbench201_comp} we compare the performance of average degree against all the baselines on NAS-Bench-201 benchmark. Our method can rank architectures effectively and outperforms the baselines on most of the datasets.

We also evaluate our method on the NDS benchmark. The result is summarized in Table \ref{table:single_metric_nds}. Similar to NASWOT \cite{mellor2021neural}, we use 1000 randomly sampled architectures. Same subset of architectures is used for all methods.


To explore the generality of the proposed \nasgraph{} framework, we examine the performance of the graph measure on micro TransNAS-Bench-101. The comparison is shown in Table \ref{table:single_metric_transbench101}. As reported in \cite{krishnakumar2022bench}, there is a pronounced variation in the ranks of training-free proxies when changing NAS benchmarks. For the \textit{class\_object} downstream task, the average degree gives the best performance. For other downstream tasks, our method also exhibits a competitive performance. We notice that other graph measure \texttt{wedge} gives a consistently higher performance compared to \texttt{avg\_deg}. The comparison of different graph measures can be found in Appendix Table \ref{table:single_metric_transbench101_surrogates}.

\subsection{Combining NASGraph metrics with data-dependent NAS metrics improves prediction}
\label{subsec_comb}

In addition to using single metric, we examine the performance of combining graph measures with existing metrics. Specifically, two metrics are combined by the summation of rankings of neural architectures by two metrics. We use a combination of \texttt{avg\_deg} and \texttt{jacob\_cov}, i.e. rank(\texttt{avg\_deg}) + rank(\texttt{jacob\_cov}). For the case of tied ranking, the average of the ranking is used.  Similar to \texttt{density} and \texttt{average degree}, the combined metrics manifest a positive correlation with test accuracy. 
By combining our metrics with \texttt{jacob\_cov}, the $\rho$ boosts to $0.85$ on CIFAR-10 and CIFAR-100 and the $\tau$ reaches $0.66$ on CIFAR-10, $0.67$ on CIFAR-100. On ImageNet-16-120, the $\rho$ increases to $0.82$ and $\tau$ is $0.64$. The comparison with existing combined metrics is shown in Table \ref{table:multiple_metric_compare}. Our combined metrics can outperform all existing methods reported in \cite{chen2021neural,abdelfattah2021zero}, which demonstrates the effectiveness of complementary nature of our proposed graph measures for NAS.
The detailed plots of combined ranks against test accuracy are shown in Appendix \ref{append:metric_comb}.

\subsection{Training-Free architecture search using NASGraph}

We evaluate the performance of our metric when used as an alternative to validation accuracy during random search. $N = 100$ and $N = 200$ architectures are randomly sampled from NAS-Bench-201 and the training-free metrics are used to perform the search. We repeat the search process 100 times, and the mean and the standard deviation are reported in Table \ref{table:nasacc_compare}. Ground truth (GT) indicates the highest validation accuracy and the highest test accuracy for the validation and the test columns respectively. It is essential to highlight the fact that all the metrics in baselines are computed based on single A40 GPU (in GPU second) while \nasgraph{} score is calculated on single CPU (in CPU second). \nasgraph{} using the surrogate model NASGraph(16, 1, 3) can outperform the other training-free metrics on CIFAR-10 and CIFAR-100 datasets with a higher mean value and a lower standard deviation. At a small cost of performance, the surrogate model NASGraph(1, 1, 3) can have a significant improvement in the computation efficiency.

\begin{table*}[hbt]
  \centering
  \caption{Comparison of training-free NAS metrics using random search method. The same subset of architectures is randomly chosen from NAS-Bench-201 for all metrics. GT reports the performance of the best architecture in that subset.}\label{table:nasacc_compare}
  \resizebox{\textwidth}{!}{%
  \begin{tabular}{cccccccccc}
    \hline
    \multirow{2}*{Metric} & \multirow{2}*{Running time} & \multicolumn{2}{c}{CIFAR-10} & & \multicolumn{2}{c}{CIFAR-100} & & \multicolumn{2}{c}{ImageNet-16-120} \\
    \cline{3-4}\cline{6-7}\cline{9-10}
     & & validation & test & & validation & test & & validation & test \\
    \hline
    \multicolumn{10}{c}{N = 100} \\
    \texttt{relu\_logdet} & $52.72$ \textcolor{DarkGray}{GPU} sec. & $89.51 \pm 0.96$ & $89.22 \pm 1.03$ & & $69.48 \pm 1.44$ & $69.58 \pm 1.50$ & & $42.92 \pm 2.41$ & $43.27 \pm 2.62$ \\  
    \texttt{grad\_norm} & $364.68$ \textcolor{DarkGray}{GPU} sec. & $88.28 \pm 1.42$ & $87.94 \pm 1.48$ & & $65.96 \pm 3.11$ & $66.13 \pm 3.10$ & & $34.97 \pm 6.82$ & $34.96 \pm 7.06$ \\
    \texttt{snip} & $363.71$ \textcolor{DarkGray}{GPU} sec. & $88.29 \pm 1.42$ & $87.95 \pm 1.48$ & & $66.14 \pm 2.96$ & $66.32 \pm 2.97$ & & $35.44 \pm 6.49$ & $35.44 \pm 6.72$ \\
    \texttt{grasp} & $377.29$ \textcolor{DarkGray}{GPU} sec. & $88.06 \pm 1.55$ & $87.74 \pm 1.58$ & & $66.27 \pm 3.50$ & $66.38 \pm 3.55$ & & $35.20 \pm 6.76$ & $35.19 \pm 6.93$ \\
    \texttt{fisher} & $315.57$ \textcolor{DarkGray}{GPU} sec. & $84.08 \pm 6.68$ & $83.70 \pm 6.67$ & & $61.77 \pm 7.26$ & $61.89 \pm 7.44$ & & $30.80 \pm 8.02$ & $30.49 \pm 8.33$ \\
    \texttt{synflow} & $360.15$ \textcolor{DarkGray}{GPU} sec. & $89.91 \pm 0.87$ & $89.67 \pm 0.88$ & & $70.03 \pm 1.79$ & $70.17 \pm 1.79$ & & $41.89 \pm 4.13$ & $42.23 \pm 4.24$ \\
    \texttt{jacob\_cov} & $360.48$ \textcolor{DarkGray}{GPU} sec. & $88.68 \pm 1.56$ & $88.32 \pm 1.59$ & & $67.45 \pm 2.91$ & $67.57 \pm 3.03$ & & $40.64 \pm 3.54$ & $40.76 \pm 3.77$ \\
    \texttt{avg\_deg} (NASGraph(1, 1, 3)) & $7.78$ \textcolor{RoyalBlue}{CPU} sec. & $89.74 \pm 0.77$ & $89.53 \pm 0.75$ & & $69.90 \pm 1.38$ & $70.01 \pm 1.43$ & & $42.00 \pm 2.80$ & $40.73 \pm 4.14$ \\
    \texttt{avg\_deg} (NASGraph(16, 1, 3)) & $106.18$ \textcolor{RoyalBlue}{CPU} sec. & $89.95 \pm 0.49$ & $89.73 \pm 0.52$ & & $70.17 \pm 1.06$ & $70.29 \pm 1.10$ & & $42.72 \pm 2.33$ & $43.15 \pm 2.29$ \\
    GT & - & $90.98 \pm 0.36$ & $90.77 \pm 0.31$ & & $71.48 \pm 0.86$ & $71.69 \pm 0.81$ & & $45.45 \pm 0.67$ & $45.74 \pm 0.65$ \\
    \hline
    \multicolumn{10}{c}{N = 200} \\
    \texttt{relu\_logdet} & $90.39$ \textcolor{DarkGray}{GPU} sec. & $89.64 \pm 0.81$ & $89.33 \pm 0.84$ & & $69.65 \pm 1.36$ & $69.87 \pm 1.33$ & & $43.25 \pm 2.22$ & $43.62 \pm 2.37$ \\
    \texttt{grad\_norm} & $644.23$ \textcolor{DarkGray}{GPU} sec. & $88.23 \pm 1.51$ & $87.87 \pm 1.53$ & & $65.46 \pm 3.34$ & $65.67 \pm 3.42$ & & $35.08 \pm 7.05$ & $35.00 \pm 7.26$ \\
    \texttt{snip} & $712.58$ \textcolor{DarkGray}{GPU} sec. & $88.23 \pm 1.51$ & $87.87 \pm 1.54$ & & $65.68 \pm 3.16$ & $65.89 \pm 3.21$ & & $35.08 \pm 7.05$ & $35.00 \pm 7.26$ \\
    \texttt{grasp} & $692.74$ \textcolor{DarkGray}{GPU} sec. & $88.31 \pm 1.35$ & $87.96 \pm 1.37$ & & $65.97 \pm 3.21$ & $66.16 \pm 3.28$ & & $34.83 \pm 6.63$ & $34.74 \pm 6.81$ \\
    \texttt{fisher} & $622.92$ \textcolor{DarkGray}{GPU} sec. & $85.55 \pm 4.91$ & $85.24 \pm 4.92$ & & $61.69 \pm 5.62$ & $61.86 \pm 5.77$ & & $29.39 \pm 6.38$ & $29.04 \pm 6.65$ \\
    \texttt{synflow} & $742.74$ \textcolor{DarkGray}{GPU} sec. & $89.87 \pm 0.85$ & $89.61 \pm 0.85$ & & $69.93 \pm 1.84$ & $70.05 \pm 1.89$ & & $41.54 \pm 3.76$ & $41.93 \pm 3.77$ \\
    \texttt{jacob\_cov} & $688.77$ \textcolor{DarkGray}{GPU} sec. & $88.34 \pm 1.67$ & $88.00 \pm 1.71$ & & $67.39 \pm 2.93$ & $67.55 \pm 3.05$ & & $40.95 \pm 3.24$ & $41.04 \pm 3.41$ \\
    \texttt{avg\_deg} (NASGraph(1, 1, 3)) & $15.98$ \textcolor{RoyalBlue}{CPU} sec. & $89.92 \pm 0.61$ & $89.69 \pm 0.62$ & & $70.25 \pm 1.20$ & $70.42 \pm 1.21$ & & $41.96 \pm 2.44$ & $42.48 \pm 2.39$ \\
    \texttt{avg\_deg} (NASGraph(16, 1, 3)) & $217.21$ \textcolor{RoyalBlue}{CPU} sec. & $89.96 \pm 0.38$ & $89.73 \pm 0.43$ & & $70.22 \pm 0.99$ & $70.45 \pm 0.98$ & & $42.27 \pm 2.36$ & $42.76 \pm 2.36$ \\
    GT & - & $91.14 \pm 0.25$ & $90.91 \pm 0.24$ & & $71.84 \pm 0.76$ & $72.04 \pm 0.72$ & & $45.72 \pm 0.54$ & $46.01 \pm 0.50$ \\
    \hline
  \end{tabular}
  }
\end{table*}

\subsection{Analysis of Operation Preference in NAS Methods}

To study the bias in architecture preference in NAS, we count the frequency (histogram) of different operations appearing in the top 10\% architectures (ranked by different metrics) on NAS-Bench-201. The visualization of the frequency distribution of operations and the details on frequency counting are given in Appendix \ref{append:op_bias}. 
The overall result is shown in Table \ref{table:freq_pref_compare}.
The ground truth (GT) ranks architecture using the test accuracy. A perfect metric is expected to have the same frequency distribution of operations as GT and thus achieve zero bias. We compute the accumulated difference in the frequency (also called bias) between a NAS metric and GT across all five operations in NAS-Bench-201.  The \texttt{avg\_deg} from \nasgraph{} has the lowest bias when using a single metric. The combined metric (\texttt{conm\_rank}) introduced in Section \ref{subsec_comb}, which is rank(\texttt{avg\_deg}) + rank(\texttt{jacob\_cov}), can further reduce the bias. The low bias towards operations can also explain why our method can have a high rank correlation in NAS via \nasgraph{}.

All of the baselines except synflow rely on a few mini-batches of data. Thus, their score varies based on the input data being used to evaluate their proxy scores. In contrast, our metric is data-agnostic and is not sensitive to the input data. Furthermore, our technique is also lightweight: it can run on a CPU and the running time is faster than most of training-free proxies running on GPU.

\section{Discussion}
In this paper, we proposed \nasgraph{}, a novel graph-based method for NAS featuring lightweight (CPU-only)  computation and is data-agnostic and training-free. Extensive experimental results verified the high correlation between the graph measure of \nasgraph{} and performance of neural architectures in several benchmarks. Compared to existing NAS methods, \nasgraph{} provides a more stable and accurate prediction of the architecture performance and can be used for efficient architecture search. We also show that our graph measures can be combined with existing data-dependent metrics to further improve NAS. We believe our findings provide a new perspective and useful tools for studying NAS through the lens of graph theory and analysis. 

Our method addresses some limitations in current NAS methods (e.g. data dependency, GPU requirement, and operation preference) and attains new state-of-the-art NAS performance.
The authors do not find any immediate ethical concerns or negative societal impacts from this study.

\bibliography{references.bib}

\begin{thebibliography}{10}

\bibitem{Zoph2018}
Barret Zoph, Vijay Vasudevan, Jonathon Shlens, and Quoc~V Le.
\newblock Learning transferable architectures for scalable image recognition.
\newblock In {\em Proceedings of the IEEE conference on computer vision and
  pattern recognition}, pages 8697--8710, 2018.

\bibitem{chen2019detnas}
Yukang Chen, Tong Yang, Xiangyu Zhang, Gaofeng Meng, Xinyu Xiao, and Jian Sun.
\newblock Detnas: Backbone search for object detection.
\newblock {\em Advances in Neural Information Processing Systems}, 32, 2019.

\bibitem{liu2019auto}
Chenxi Liu, Liang-Chieh Chen, Florian Schroff, Hartwig Adam, Wei Hua, Alan~L
  Yuille, and Li~Fei-Fei.
\newblock Auto-deeplab: Hierarchical neural architecture search for semantic
  image segmentation.
\newblock In {\em Proceedings of the IEEE/CVF conference on computer vision and
  pattern recognition}, pages 82--92, 2019.

\bibitem{zoph2016neural}
Barret Zoph and Quoc~V Le.
\newblock Neural architecture search with reinforcement learning.
\newblock {\em arXiv preprint arXiv:1611.01578}, 2016.

\bibitem{Pham2018}
Hieu Pham, Melody Guan, Barret Zoph, Quoc Le, and Jeff Dean.
\newblock Efficient neural architecture search via parameters sharing.
\newblock In {\em International conference on machine learning}, pages
  4095--4104. PMLR, 2018.

\bibitem{chu2021fairnas}
Xiangxiang Chu, Bo~Zhang, and Ruijun Xu.
\newblock Fairnas: Rethinking evaluation fairness of weight sharing neural
  architecture search.
\newblock In {\em Proceedings of the IEEE/CVF International Conference on
  computer vision}, pages 12239--12248, 2021.

\bibitem{Liu2018}
Chenxi Liu, Barret Zoph, Maxim Neumann, Jonathon Shlens, Wei Hua, Li-Jia Li,
  Li~Fei-Fei, Alan Yuille, Jonathan Huang, and Kevin Murphy.
\newblock Progressive neural architecture search.
\newblock In {\em Proceedings of the European conference on computer vision
  (ECCV)}, pages 19--34, 2018.

\bibitem{liu2018darts}
Hanxiao Liu, Karen Simonyan, and Yiming Yang.
\newblock Darts: Differentiable architecture search.
\newblock {\em arXiv preprint arXiv:1806.09055}, 2018.

\bibitem{white2021bananas}
Colin White, Willie Neiswanger, and Yash Savani.
\newblock Bananas: Bayesian optimization with neural architectures for neural
  architecture search.
\newblock In {\em Proceedings of the AAAI Conference on Artificial
  Intelligence}, volume~35 of {\em 12}, pages 10293--10301, 2021.

\bibitem{mellor2021neural}
Joe Mellor, Jack Turner, Amos Storkey, and Elliot~J Crowley.
\newblock Neural architecture search without training.
\newblock In {\em International Conference on Machine Learning}, pages
  7588--7598. PMLR, 2021.

\bibitem{abdelfattah2021zero}
Mohamed~S Abdelfattah, Abhinav Mehrotra, {\L}ukasz Dudziak, and Nicholas~D
  Lane.
\newblock Zero-cost proxies for lightweight nas.
\newblock {\em arXiv preprint arXiv:2101.08134}, 2021.

\bibitem{xu2019pc}
Yuhui Xu, Lingxi Xie, Xiaopeng Zhang, Xin Chen, Guo-Jun Qi, Qi~Tian, and
  Hongkai Xiong.
\newblock Pc-darts: Partial channel connections for memory-efficient
  architecture search.
\newblock {\em arXiv preprint arXiv:1907.05737}, 2019.

\bibitem{wang2021_dartspt}
Ruochen Wang, Minhao Cheng, Xiangning Chen, Xiaocheng Tang, and Cho{-}Jui
  Hsieh.
\newblock Rethinking architecture selection in differentiable {NAS}.
\newblock In {\em 9th International Conference on Learning Representations,
  {ICLR} 2021, Virtual Event, Austria, May 3-7, 2021}. OpenReview.net, 2021.

\bibitem{zela2020_robust_darts}
Arber Zela, Thomas Elsken, Tonmoy Saikia, Yassine Marrakchi, Thomas Brox, and
  Frank Hutter.
\newblock Understanding and robustifying differentiable architecture search.
\newblock In {\em 8th International Conference on Learning Representations,
  {ICLR} 2020, Addis Ababa, Ethiopia, April 26-30, 2020}. OpenReview.net, 2020.

\bibitem{yu2020_Evaluating}
Kaicheng Yu, Christian Sciuto, Martin Jaggi, Claudiu Musat, and Mathieu
  Salzmann.
\newblock Evaluating the search phase of neural architecture search.
\newblock In {\em 8th International Conference on Learning Representations,
  {ICLR} 2020, Addis Ababa, Ethiopia, April 26-30, 2020}. OpenReview.net, 2020.

\bibitem{Guo2020_spos}
Zichao Guo, Xiangyu Zhang, Haoyuan Mu, Wen Heng, Zechun Liu, Yichen Wei, and
  Jian Sun.
\newblock Single path one-shot neural architecture search with uniform
  sampling.
\newblock In Andrea Vedaldi, Horst Bischof, Thomas Brox, and Jan{-}Michael
  Frahm, editors, {\em Computer Vision - {ECCV} 2020 - 16th European
  Conference, Glasgow, UK, August 23-28, 2020, Proceedings, Part {XVI}}, volume
  12361 of {\em Lecture Notes in Computer Science}, pages 544--560. Springer,
  2020.

\bibitem{Bender2018_understanding}
Gabriel Bender, Pieter{-}Jan Kindermans, Barret Zoph, Vijay Vasudevan, and
  Quoc~V. Le.
\newblock Understanding and simplifying one-shot architecture search.
\newblock In Jennifer~G. Dy and Andreas Krause, editors, {\em Proceedings of
  the 35th International Conference on Machine Learning, {ICML} 2018,
  Stockholmsm{\"{a}}ssan, Stockholm, Sweden, July 10-15, 2018}, volume~80 of
  {\em Proceedings of Machine Learning Research}, pages 549--558. {PMLR}, 2018.

\bibitem{Noy2020_asap}
Asaf Noy, Niv Nayman, Tal Ridnik, Nadav Zamir, Sivan Doveh, Itamar Friedman,
  Raja Giryes, and Lihi Zelnik.
\newblock {ASAP:} architecture search, anneal and prune.
\newblock In Silvia Chiappa and Roberto Calandra, editors, {\em The 23rd
  International Conference on Artificial Intelligence and Statistics, {AISTATS}
  2020, 26-28 August 2020, Online [Palermo, Sicily, Italy]}, volume 108 of {\em
  Proceedings of Machine Learning Research}, pages 493--503. {PMLR}, 2020.

\bibitem{zhou2020econas}
Dongzhan Zhou, Xinchi Zhou, Wenwei Zhang, Chen~Change Loy, Shuai Yi, Xuesen
  Zhang, and Wanli Ouyang.
\newblock Econas: Finding proxies for economical neural architecture search.
\newblock In {\em Proceedings of the IEEE/CVF Conference on computer vision and
  pattern recognition}, pages 11396--11404, 2020.

\bibitem{hanin2019complexity}
Boris Hanin and David Rolnick.
\newblock Complexity of linear regions in deep networks.
\newblock In {\em International Conference on Machine Learning}, pages
  2596--2604. PMLR, 2019.

\bibitem{chen2021neural}
Wuyang Chen, Xinyu Gong, and Zhangyang Wang.
\newblock Neural architecture search on imagenet in four gpu hours: A
  theoretically inspired perspective.
\newblock {\em arXiv preprint arXiv:2102.11535}, 2021.

\bibitem{krishnakumar2022bench}
Arjun Krishnakumar, Colin White, Arber Zela, Renbo Tu, Mahmoud Safari, and
  Frank Hutter.
\newblock Nas-bench-suite-zero: Accelerating research on zero cost proxies.
\newblock {\em arXiv preprint arXiv:2210.03230}, 2022.

\bibitem{verdenius2020pruning}
Stijn Verdenius, Maarten Stol, and Patrick Forr{\'e}.
\newblock Pruning via iterative ranking of sensitivity statistics.
\newblock {\em arXiv preprint arXiv:2006.00896}, 2020.

\bibitem{tanaka2020pruning}
Hidenori Tanaka, Daniel Kunin, Daniel~L Yamins, and Surya Ganguli.
\newblock Pruning neural networks without any data by iteratively conserving
  synaptic flow.
\newblock {\em Advances in neural information processing systems},
  33:6377--6389, 2020.

\bibitem{ying2019bench}
Chris Ying, Aaron Klein, Eric Christiansen, Esteban Real, Kevin Murphy, and
  Frank Hutter.
\newblock Nas-bench-101: Towards reproducible neural architecture search.
\newblock In {\em International Conference on Machine Learning}, pages
  7105--7114. PMLR, 2019.

\bibitem{dong2020bench}
Xuanyi Dong and Yi~Yang.
\newblock Nas-bench-201: Extending the scope of reproducible neural
  architecture search.
\newblock {\em arXiv preprint arXiv:2001.00326}, 2020.

\bibitem{duan2021transnas}
Yawen Duan, Xin Chen, Hang Xu, Zewei Chen, Xiaodan Liang, Tong Zhang, and
  Zhenguo Li.
\newblock Transnas-bench-101: Improving transferability and generalizability of
  cross-task neural architecture search.
\newblock In {\em Proceedings of the IEEE/CVF Conference on Computer Vision and
  Pattern Recognition}, pages 5251--5260, 2021.

\bibitem{radosavovic2019network}
Ilija Radosavovic, Justin Johnson, Saining Xie, Wan-Yen Lo, and Piotr
  Doll{\'a}r.
\newblock On network design spaces for visual recognition.
\newblock In {\em Proceedings of the IEEE/CVF international conference on
  computer vision}, pages 1882--1890, 2019.

\bibitem{Krizhevsky2009}
Alex Krizhevsky, Geoffrey Hinton, et~al.
\newblock Learning multiple layers of features from tiny images.
\newblock Technical report, Univeristy of Toronto, 2009.

\bibitem{chrabaszcz2017downsampled}
Patryk Chrabaszcz, Ilya Loshchilov, and Frank Hutter.
\newblock A downsampled variant of imagenet as an alternative to the cifar
  datasets.
\newblock {\em arXiv preprint arXiv:1707.08819}, 2017.

\bibitem{zamir2018taskonomy}
Amir~R Zamir, Alexander Sax, William Shen, Leonidas~J Guibas, Jitendra Malik,
  and Silvio Savarese.
\newblock Taskonomy: Disentangling task transfer learning.
\newblock In {\em Proceedings of the IEEE conference on computer vision and
  pattern recognition}, pages 3712--3722, 2018.

\bibitem{real2019regularized}
Esteban Real, Alok Aggarwal, Yanping Huang, and Quoc~V Le.
\newblock Regularized evolution for image classifier architecture search.
\newblock In {\em Proceedings of the aaai conference on artificial
  intelligence}, volume~33, pages 4780--4789, 2019.

\bibitem{pham2018efficient}
Hieu Pham, Melody Guan, Barret Zoph, Quoc Le, and Jeff Dean.
\newblock Efficient neural architecture search via parameters sharing.
\newblock In {\em International conference on machine learning}, pages
  4095--4104. PMLR, 2018.

\bibitem{liu2018progressive}
Chenxi Liu, Barret Zoph, Maxim Neumann, Jonathon Shlens, Wei Hua, Li-Jia Li,
  Li~Fei-Fei, Alan Yuille, Jonathan Huang, and Kevin Murphy.
\newblock Progressive neural architecture search.
\newblock In {\em Proceedings of the European conference on computer vision
  (ECCV)}, pages 19--34, 2018.

\bibitem{li2023zico}
Guihong Li, Yuedong Yang, Kartikeya Bhardwaj, and Radu Marculescu.
\newblock Zico: Zero-shot nas via inverse coefficient of variation on
  gradients.
\newblock {\em arXiv preprint arXiv:2301.11300}, 2023.

\bibitem{du2018gradient}
Simon~S Du, Xiyu Zhai, Barnabas Poczos, and Aarti Singh.
\newblock Gradient descent provably optimizes over-parameterized neural
  networks.
\newblock {\em arXiv preprint arXiv:1810.02054}, 2018.

\bibitem{lee2018snip}
Namhoon Lee, Thalaiyasingam Ajanthan, and Philip~HS Torr.
\newblock Snip: Single-shot network pruning based on connection sensitivity.
\newblock {\em arXiv preprint arXiv:1810.02340}, 2018.

\bibitem{wang2020picking}
Chaoqi Wang, Guodong Zhang, and Roger Grosse.
\newblock Picking winning tickets before training by preserving gradient flow.
\newblock {\em arXiv preprint arXiv:2002.07376}, 2020.

\bibitem{theis2018faster}
Lucas Theis, Iryna Korshunova, Alykhan Tejani, and Ferenc Husz{\'a}r.
\newblock Faster gaze prediction with dense networks and fisher pruning.
\newblock {\em arXiv preprint arXiv:1801.05787}, 2018.

\bibitem{liu2022convnet}
Zhuang Liu, Hanzi Mao, Chao-Yuan Wu, Christoph Feichtenhofer, Trevor Darrell,
  and Saining Xie.
\newblock A convnet for the 2020s.
\newblock In {\em Proceedings of the IEEE/CVF Conference on Computer Vision and
  Pattern Recognition}, pages 11976--11986, 2022.

\bibitem{gao2016universal}
Jianxi Gao, Baruch Barzel, and Albert-L{\'a}szl{\'o} Barab{\'a}si.
\newblock Universal resilience patterns in complex networks.
\newblock {\em Nature}, 530(7590):307--312, 2016.

\bibitem{gupta2014decompositions}
Rishi Gupta, Tim Roughgarden, and Comandur Seshadhri.
\newblock Decompositions of triangle-dense graphs.
\newblock In {\em Proceedings of the 5th conference on Innovations in
  theoretical computer science}, pages 471--482, 2014.

\end{thebibliography}
\bibliographystyle{unsrt}

\newpage
\appendix
\section{Appendix}


\subsection{Proof of Theorem \ref{th:conv_relu}}
\label{append:conv_relu}

\begin{reptheorem}{th:conv_relu}
Let $\mathbf{x}$ be the input to $\mathcal{F}$. $\mathcal{F}$ is converted to a graph $G(V, E)$ using NASGraph framework: the input to $\mathcal{F}$ has $\mathds{1} \in \mathbb{R}^{H^{(0)} \times W^{(0)}}$ for the channel $i$ and $\mathds{O} \in \mathbb{R}^{H^{(0)} \times W^{(0)}}$ for rest of channels. There is no edge between node i and node j if and only if the output for the channel j is all-zeros matrix $\mathds{O} \in \mathbb{R}^{H^{(1)} \times W^{(1)}}$.
\end{reptheorem}

\textbf{Proof:} $y_i^{(1)} = \mathcal{F}(\mathbf{x})$, and $(y_i^{(1)})_{d_1d_2d_3} \in \mathbb{R}$ is computed by:

\begin{equation}
    \begin{aligned}
    & (y_i^{(1)})_{d_1d_2d_3} = \rm{ReLU}( \\
    & \sum_{a=0}^{f_{k_1}^{(0)}}\sum_{b=0}^{f_{k_2}^{(0)}}\sum_{c=0}^{C^{(0)}} \mathbf{w}_{a,b,c}^{(0),d_3}\mathbf{x}^{(0)}_{a+(d_1-1)s_l, b+(d_2-1)s_l,c} + \mathbf{b}^{(0),d_3}) \\
    \end{aligned}
\end{equation}

where $f^{(0)}_{k_1}$ is the kernel height, $f^{(0)}_{k_2}$ is the kernel width, and $s_l$ is the convolution stride. Since $(y_i^{(1)})_{d_1d_2d_3} \ge 0$, according to Equation \ref{eq:agg}, $\omega_{i^{(l-1)}j^{(l)}} = 0$ implies $(y_i^{(1)})_{d_1d_2d_3} = 0$. The reverse implication is obvious: when $(y_i^{(1)})_{d_1d_2d_3} = 0, \; \forall d_1, d_2 \text{ s.t. } 0 \le d_1 \le H^{(1)}, 0 \le d_2 \le W^{(1)}$, then the summation $\omega_{i^{(l-1)}j^{(l)}} = 0$. This completes our proof.


\subsection{Bias in NAS methods}
\label{append:op_bias}

\begin{figure}[hbt]
  \includegraphics[width=\linewidth]{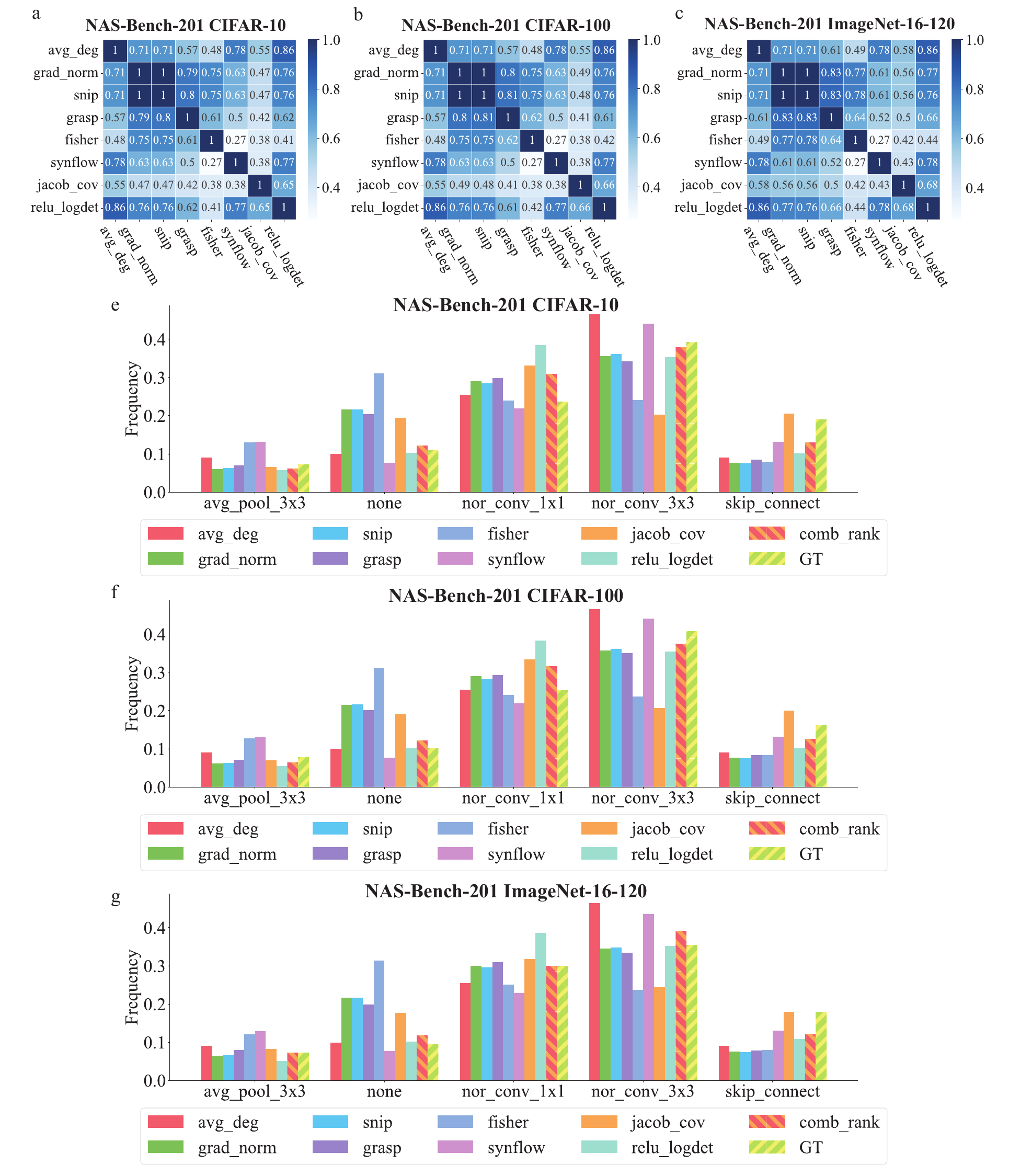}
  \caption{(a)-(c) Spearman's ranking correlation $\rho$ between pair of metrics and (e)-(f) the preference for different operations on NAS-Bench-201. GT is the operation distribution of top architectures ranked by test accuracy.}
  \label{fig:metric_op_compare}
\end{figure}

We rank architectures in NAS-Bench-201 using different metrics and compute the $\rho$ of architecture ranks for each pair of metrics. Figure \ref{fig:metric_op_compare} (a)-(c) show the pairwise ranking correlation between metrics. Both \texttt{avg\_deg} and \texttt{synflow} are data-agnostic approach and the correlation between these two metrics is high. Besides, the correlations between these two metrics do no change across datasets. The correlation between \texttt{avg\_deg} and \texttt{jacob\_cov} is comparably small, but a combination of these two metrics can have the best ranking correlation between the metrics and the performance of neural architectures. Because of rounding number, \texttt{grad\_norm} and \texttt{snip} have the $\rho$ of 1. But they do not give the exactly same rankings of neural architectures. When we check the architecture rankings precisely using these two metrics, we find $\rho = 0.9982$ on CIFAR-10, $\rho = 0.9984$ on CIFAR-100 and $\rho = 0.9988$ on ImageNet-16-120.

We extract top 10\% neural architectures on NAS-Bench-201 and count the frequency of each operation (\texttt{avg\_pool}, \texttt{none}, \texttt{nor\_conv\_1$\times$1}, \texttt{nor\_conv\_3$\times$3}, \texttt{skip\_connect}) appearing in the selected subset. Figure \ref{fig:metric_op_compare} (e)-(f) show the frequency of different operations on different datasets. There is no pronounced difference in the distribution across different datasets for GT. This is reasonable when we examine the distribution of test accuracy in CIFAR-10, CIFAR-100 and ImageNet-16-120 datasets (Figure \ref{fig:nasbenchmark_corr} (f)) where test accuracy is positively correlated among datasets. These facts indicate the same architecture generally have a similar performance on different datasets. Therefore, the data-agnostic NAS methods (\texttt{avg\_deg} and \texttt{synflow}) can be effective to search neural architectures across different datasets.

Based on the distribution of operation preference, we find our metric \texttt{avg\_deg} has a distribution similar to the GT. Similar to \texttt{synflow}, our metric has a relatively low preference for \texttt{skip\_connect} while \texttt{jacob\_cov} has a high preference for \texttt{skip\_connect}. That might explain the reason why a combination of \texttt{avg\_deg} and \texttt{jacob\_cov} gives the highest correlation between test accuracy and combined metrics. Another reasoning is related to the definition of \texttt{jacob\_cov}. The Jacobian for the $i$-th neuron in the output of the layer $L$ with parameter $\theta_{\alpha}$ valuated at a point \textbf{x} is defined as \cite{mellor2021neural}:

\begin{equation}
  J_{i\alpha}(\mathbf{x}) = \partial_{\mathbf{\theta}_{\alpha}}z^{(L)}_{i}(\mathbf{x})
\end{equation}

The \texttt{jacob\_cov} metric takes gradient of model parameters into consideration, i.e., it focuses on the backward propagation process. The \nasgraph{} framework, on the other hand, considers the forward propagation for each graph block. They are complementary to each other. Hence, Combining \texttt{avg\_deg} with \texttt{jacob\_cov} leads to a higher ranking correlation. \texttt{comb\_rank} is rank(\texttt{avg\_deg}) + rank(\texttt{jacob\_cov}). The frequency distribution of \texttt{comb\_rank} is very close to \texttt{GT} as shown in Figure \ref{fig:metric_op_compare} (e)-(g).

The \texttt{relu\_logdet} metric exhibits the highest preference for convolution with small kernel size while GT indicates the best neural architectures prefer convolution with large kernel size. \texttt{relu\_logdet} might have a problem when the search space $\mathcal{A}$ includes convolution with large kernel size. ConvNeXt \cite{liu2022convnet} has shown the advantage of modern design using convolution with large kernel size. \texttt{fisher} has the highest preference for \texttt{none} operation, and it has the lowest correlation in the NAS-Bench-201.

\subsection{Output combination from preceding graph blocks}
\label{append:sum_concate}

\begin{figure}[hbt]
  \includegraphics[width=\linewidth]{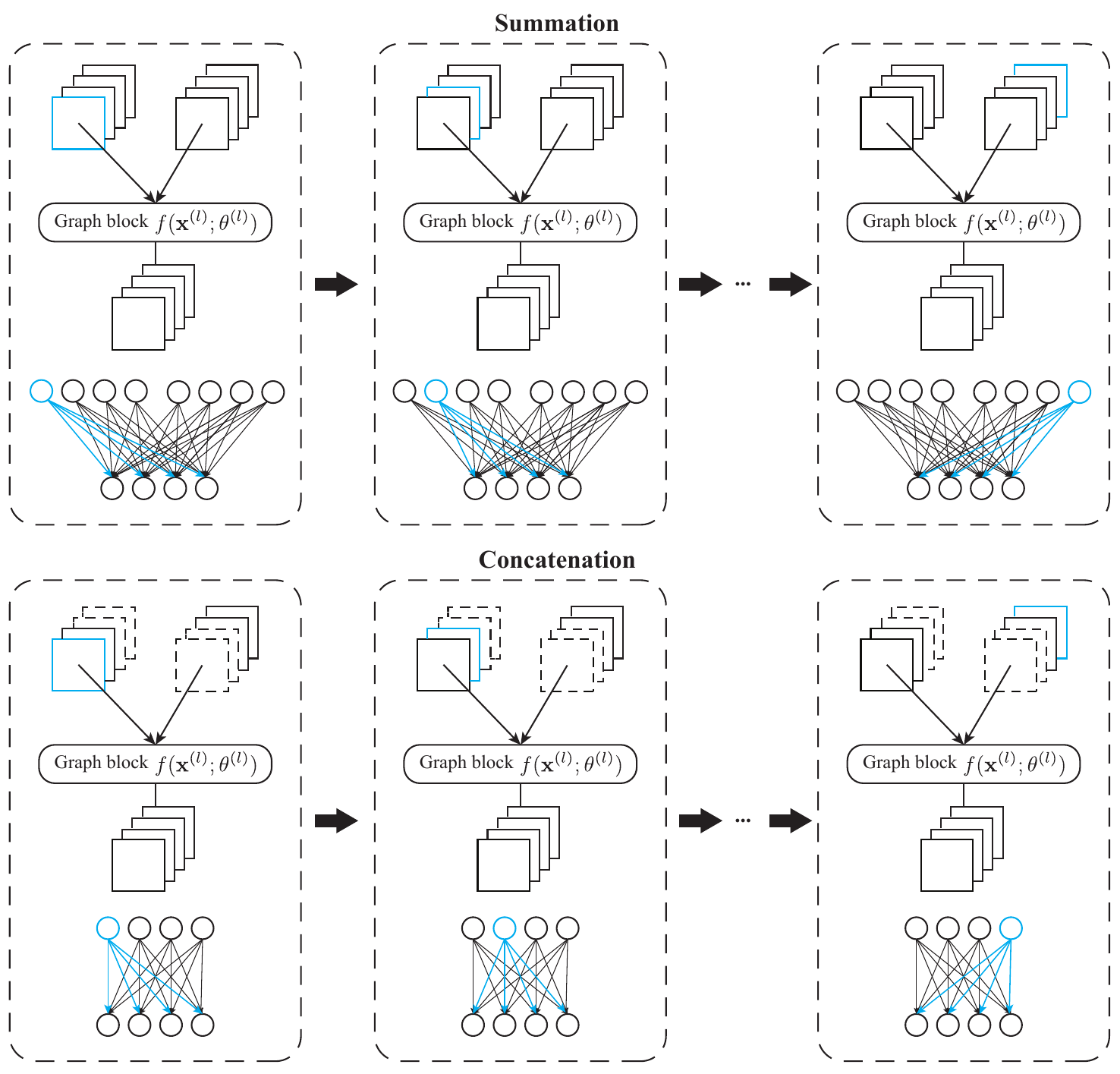}
  \caption{An illustration of converting one graph block of the neural architecture to subgraph in the case of summation and concatenation. Channels in dashed line mean virtually concatenated channels. These channels become zero matrix after applying a mask. Connection is established if the score between two graph nodes is non-zero.}
  \label{fig:agg_method}
\end{figure}

When there are multiple graph blocks connected to the same graph block, there are two ways to combine the outputs from preceding graph blocks: \textit{summation} and \textit{concatenation}, as illustrated in the middle of Figure \ref{fig:supernode_agg}. In the case of summation, we do the forward propagation for each channel of all branches. In the case of concatenation, however, the outputs of the preceding graph blocks do not match the input dimension of the current graph block. We add virtual channels to the outputs of the preceding graph blocks such that the output dimension matches the input dimension, and hence we can do the conversion in the block-wise fashion. We want to emphasize that the summation or concatenation is determined by the original neural architecture. The graph block just combines components in the neural architecture such as Conv and ReLU. $w.l.o.g.$ we examine the case of concatenation and the case of summation of two preceding graph blocks connecting to the $l$-th graph block as shown in Figure \ref{fig:agg_method}. In the case of summation, two graph blocks connect to the $l$-th graph block. The output of each graph block has 4 channels. The input of $l$-th graph block has 4 channels. In the case of concatenation, two outputs from the preceding graph blocks have 2 channels while the input of $l$-th graph block has 4 channels.

\textbf{Summation} The summation requires that outputs of the preceding graph blocks have the same dimension as the input of the current graph block. We perform forward propagation for each channel of the outputs of the preceding graph blocks. Hence, there are 8 forward propagations in the case shown in Figure \ref{fig:agg_method}. Each forward propagation determines the scores between the ``activated'' channel and all the output channels of $l$-th graph block.

\textbf{Concatenation} The concatenation requires that outputs of the preceding graph blocks have the same dimension as the input of the current graph block except for the channel dimension, and the summation of the channel size of the outputs of the preceding graph blocks equal to the channel size of the input of the current graph block. Because we do the conversion in the block-wise fashion, there will be dimension mismatch if we do forward propagation using the dimension of the preceding outputs. So we add virtual channels to make sure two dimensions match. The number of added channels is $C^{(l)} - C^{(l-1)}$. In the case of Figure \ref{fig:agg_method}, two virtual channels are added for each input. We do not activate ``virtual channels'' in any forward propagation, so there are 4 forward propagations in total.

\subsection{Parallel computing for graph blocks}

The direct way to compute the weight of a graph edge is to do forward propagation as many times as the number of input channels, and in this case, the batch size of the input is one. In other words, to compute the score $\omega_{i^{(l-1)}j^{(l)}}$, the input $\mathcal{M}_{c}^{(l)} \odot x^{(l)}$ has the batch dimension of one. To enable parallel computing of scores, we concatenate inputs for the same graph block on the batch dimension $[\mathcal{M}_{1}^{(l)} \odot x^{(l)}, \ldots, \mathcal{M}_{c}^{(l)} \odot x^{(l)}]$ so that scores can be computed independently and in parallel. The batch size, under this condition, is equal to the input channel size. Using this approach, the batch size dimension has a different meaning compared to the standard definition in batch normalization. Moreover, the effective batch dimension is essentially one as we only use the same input $\mathbf{x}^{(l)}$ to determine scores. Therefore,  during graph conversion, we remove batch normalization in the entire neural architecture. For example, Conv-BN-ReLU becomes Conv-ReLU.

\subsection{Albation Study}
\label{append:ablation}

\begin{figure}[hbt]
  \includegraphics[width=\linewidth]{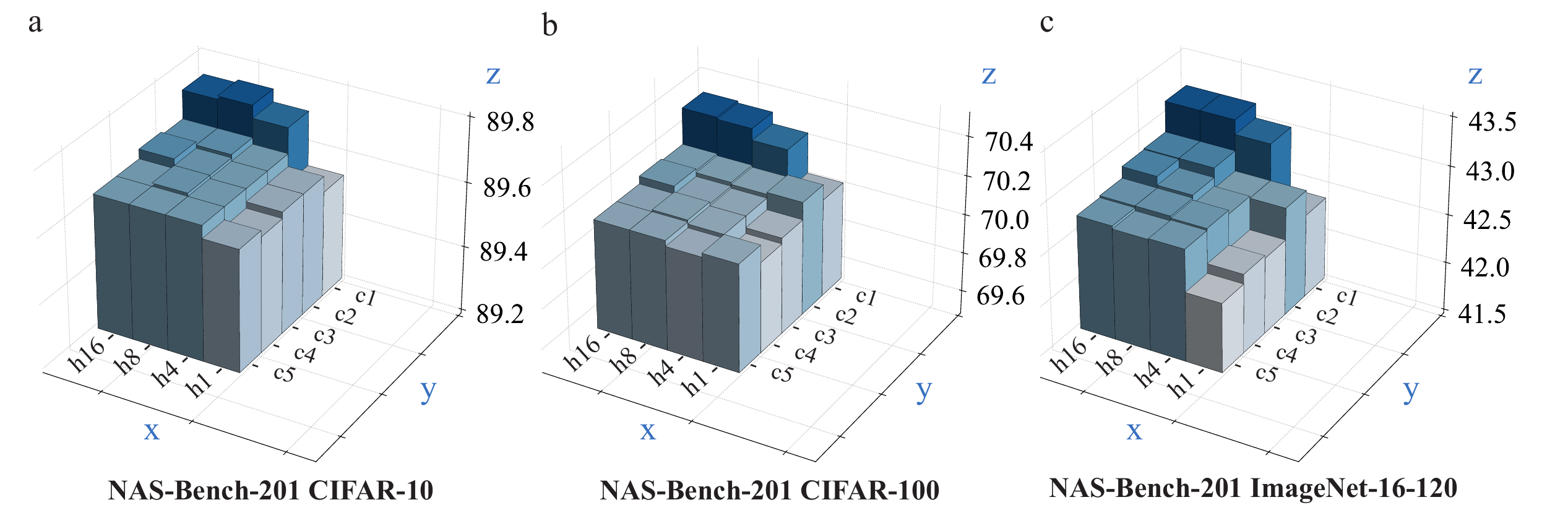}
  \caption{3D-Bar chart for the reducing factors. The number of channels $h$ and number of cells $c$ within a module change in the ablation study. We examine 4 different number of channels along the $x$ axis: $h \in \{1, 4, 8, 10\}$ and 5 different number of cells along the $y$ axis: $c \in \{1, 2, 3, 4, 5\}$. The $z$ axis is the test accuracies of neural architectures.}
  \label{fig:ablation_3d}
\end{figure}

We use ablation study to analyze the effect of using surrogate models. There are two reducing factors: (1) number of channels, (2) number of search cells within one module. The random search result on NAS-Bench-201 is listed in Table \ref{table:nasacc_compare_full} for the case $N = 100$ and $N = 200$. GT reports the highest test accuracy of neural architectures within the selected subset in the random search process. In both $N = 100$ and $N = 200$ cases, we do not find a significant variation of the performance of surrogate models except for the efficiency. As we decrease the number of cells or number of channels, there is a significant improvement in the efficiency. We use a grid routine to systematically study the effect of number of channels and number of cells on the performance of the surrogate model. Figure \ref{fig:ablation_3d} shows the effect of varying number of channels and number of cells within the same module on NAS-Bench-201 using random search algorithm. The number of sampled architectures is 100 and each search process is repeated for 100 times. 

\textbf{Number of channels} Along $x$ axis in Figure \ref{fig:ablation_3d} shows the effect of varying the number of channels. We find as the number of channels increase, the performance of the surrogate models either does not change or improves. This is expected because decreasing the number of channels make the equivalent neural architecture has less model complexity (number of model parameters decrease) compared to the original one. 

\textbf{Number of cells} Along $y$ axis in Figure \ref{fig:ablation_3d} shows the effect of changing the number of cells within a module. We find there is no monotonous increase nor decrease in the performance as the number of cells varies. We believe it is related to the way we bridge neural architecture space and graph space. Because we convert neural architectures in the block-wise fashion and input is independent of other graph blocks, the conversion of one graph block is independent of other graph blocks. Besides, the cell structure is the same within the same module. Considering the cell structure that corresponds to a subgraph, the module consisting of a linear stack of cell structures corresponds to a stack of identical subgraphs. Therefore, graph measures using surrogate models that decrease the number of cells is expected to not impose a significant effect on the graph measures, and hence rankings of the original neural network models.

Overall, we find that there is no remarkable difference in the accuracies using the random search algorithm. However, we can vastly boost the efficiency by the means of surrogate models. We choose the surrogate model \nasgraph{(16, 1, 3)} to compute graph measures on different NAS benchamrks.

\begin{table*}[hbt]
  \centering
  \caption{Comparison of different surrogate models on NAS-Bench-201. \texttt{avg\_deg} is used as the metric to score architectures. Reported results are averaged over 100 runs, and both mean values and standard deviations are recorded. GT records the highest accuracies of the randomly sampled architectures.}\label{table:nasacc_compare_full}
  \vspace{1em}
  \resizebox{\textwidth}{!}{%
  \begin{tabular}{cccccccccc}
    \hline
    \multirow{2}*{Method} & \multirow{2}*{Time} & \multicolumn{2}{c}{CIFAR-10} & & \multicolumn{2}{c}{CIFAR-100} & & \multicolumn{2}{c}{ImageNet-16-120} \\
    \cline{3-4}\cline{6-7}\cline{9-10}
     & & validation & test & & validation & test & & validation & test \\
    \hline
    \multicolumn{10}{c}{N = 100} \\
    \nasgraph{(1, 1, 3)} & $7.78$ sec. & $89.74 \pm 0.77$ & $89.53 \pm 0.75$ & & $69.90 \pm 1.38$ & $70.01 \pm 1.43$ & & $42.00 \pm 2.80$ & $40.73 \pm 4.14$ \\
    \nasgraph{(4, 1, 3)} & $19.23$ sec. & $89.91 \pm 0.49$ & $89.70 \pm 0.52$ & & $70.05 \pm 1.16$ & $70.22 \pm 1.18$ & & $42.60 \pm 2.43$ & $43.00 \pm 2.42$ \\
    \nasgraph{(8, 1, 3)} & $62.61$ sec. & $89.96 \pm 0.46$ & $89.74 \pm 0.49$ & & $70.15 \pm 1.01$ & $70.28 \pm 1.05$ & & $42.72 \pm 2.33$ & $43.15 \pm 2.30$ \\
    \nasgraph{(16, 1, 3)} & $106.18$ sec. & $89.95 \pm 0.49$ & $89.73 \pm 0.52$ & & $70.17 \pm 1.06$ & $70.29 \pm 1.10$ & & $42.72 \pm 2.33$ & $43.15 \pm 2.29$ \\
    \nasgraph{(1, 5, 3)} & $25.63$ sec. & $89.78 \pm 0.64$ & $89.58 \pm 0.65$ & & $69.93 \pm 1.23$ & $70.06 \pm 1.30$ & & $41.80 \pm 2.66$ & $42.22 \pm 2.63$ \\
    \nasgraph{(4, 5, 3)} & $89.68$ sec. & $89.84 \pm 0.53$ & $89.62 \pm 0.55$ & & $69.88 \pm 1.07$ & $70.04 \pm 1.13$ & & $42.22 \pm 2.56$ & $42.66 \pm 2.56$ \\
    \nasgraph{(8, 5, 3)} & $258.03$ sec. & $89.84 \pm 0.52$ & $89.62 \pm 0.55$ & &$69.90 \pm 1.07$ & $70.07 \pm 1.14$ & & $42.18 \pm 2.57$ & $42.65 \pm 2.57$ \\
    GT & - & $90.98 \pm 0.36$ & $90.77 \pm 0.31$ & & $71.48 \pm 0.86$ & $71.69 \pm 0.81$ & & $45.45 \pm 0.67$ & $45.74 \pm 0.65$ \\
    \hline
    \multicolumn{10}{c}{N = 200} \\
    \nasgraph{(1, 1, 3)} & $15.98$ sec. & $89.92 \pm 0.61$ & $89.69 \pm 0.62$ & & $70.25 \pm 1.20$ & $70.42 \pm 1.21$ & & $41.96 \pm 2.44$ & $42.48 \pm 2.39$ \\
    \nasgraph{(4, 1, 3)} & $61.60$ sec. & $89.95 \pm 0.38$ & $89.73 \pm 0.42$ & & $70.17 \pm 1.01$ & $70.42 \pm 0.99$ & & $42.25 \pm 2.33$ & $42.73 \pm 2.32$ \\
    \nasgraph{(8, 1, 3)} & $186.21$ sec. & $89.97 \pm 0.37$ & $89.74 \pm 0.42$ & & $70.25 \pm 0.89$ & $70.47 \pm 0.87$ & & $42.40 \pm 2.11$ & $42.87 \pm 2.13$ \\
    \nasgraph{(16, 1, 3)} & $217.21$ sec. & $89.96 \pm 0.38$ & $89.73 \pm 0.43$ & & $70.22 \pm 0.99$ & $70.45 \pm 0.98$ & & $42.27 \pm 2.36$ & $42.76 \pm 2.36$ \\
    \nasgraph{(1, 5, 3)} & $50.33$ sec. & $89.80 \pm 0.59$ & $89.61 \pm 0.59$ & & $70.05 \pm 1.15$ & $70.18 \pm 1.23$ & & $41.47 \pm 2.50$ & $41.97 \pm 2.51$ \\
    \nasgraph{(4, 5, 3)} & $196.08$ sec. & $89.89 \pm 0.45$ & $89.68 \pm 0.49$ & & $70.05 \pm 0.93$ & $70.21 \pm 1.05$ & & $41.91 \pm 2.04$ & $42.45 \pm 2.05$ \\
    \nasgraph{(8, 5, 3)} & $413.47$ sec. & $89.88 \pm 0.45$ & $89.66 \pm 0.49$ & & $70.05 \pm 0.95$ & $70.23 \pm 1.07$ & & $42.01 \pm 2.07$ & $42.53 \pm 2.09$ \\
    GT & - & $91.14 \pm 0.25$ & $90.91 \pm 0.24$ & & $71.84 \pm 0.76$ & $72.04 \pm 0.72$ & & $45.72 \pm 0.54$ & $46.01 \pm 0.50$ \\
    \hline
  \end{tabular}
  }
\end{table*}

\subsection{Variation in Rankings Using Different Parameter Initialization}
\label{append:rank_var}

\begin{figure}[hbt]
  \centering
  \includegraphics[width=0.56\linewidth]{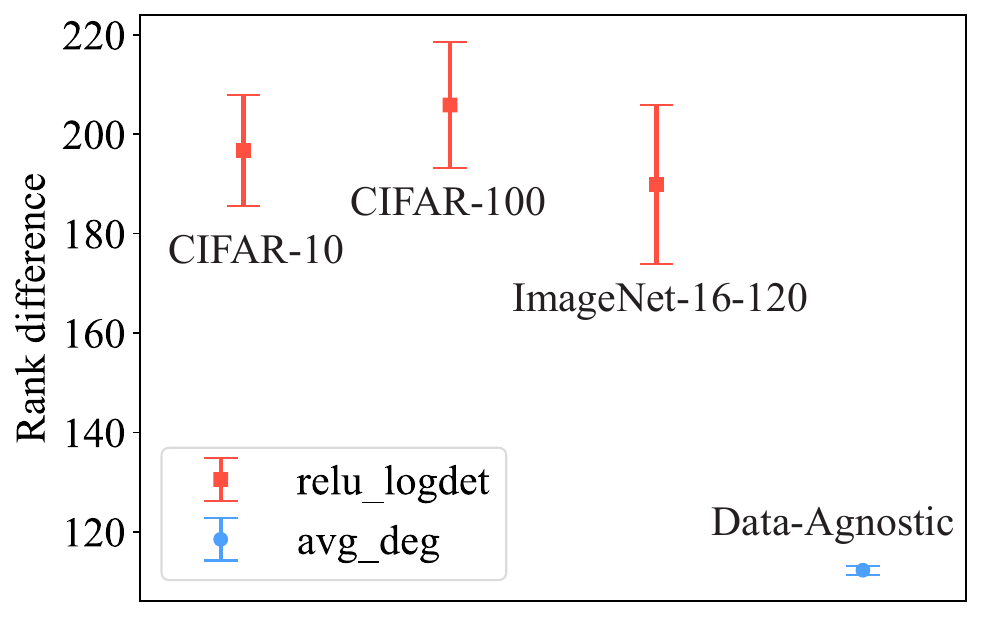}
  \caption{The variation of the architecture rankings using \texttt{relu\_logdet} and \texttt{avg\_deg} as the ranking metric. The mean value and standard deviation is calculated over 8 random seeds.}
  \label{fig:nas201metricvar}
\end{figure}

Because the training-free NAS methods do a single forward/backward propagation on models with randomly initialized parameters, it is potentially subjected to different random initialization. To examine the variation in rankings due to different initialization of model parameters, we repeat the computation of metrics for 8 times. Each time, a different initialization is used, and the initialization follows normal distribution. We use the pair rank difference to indicate the variation in the rankings for a pair of two random processes. The pair rank difference is defined by:

\begin{equation}
  \rm{pair \; rank \;  difference} = \sum_{k=0}^{n} |\rm{rank}_{i}(a_{k}) - \rm{rank}_{j}(a_{k})|
\end{equation}

where $n$ is the total number of neural architectures, on NAS-Bench-201 benchmark, $n = 15,625$. $\rm{rank}_{i}(a_k)$ and $\rm{rank}_{j}(a_{k})$ are the ranking of $k$-th architecture in the $i$-th and $j$-th random initialization processes, respectively. Considering the computational overheads, we choose NASWOT, a NAS method requires training dataset, to compute the variation in rankings in comparison with our method. We use 8 random seeds and compute the pair rank difference. Figure \ref{fig:nas201metricvar} shows the pair rank difference on NAS-Bench-201 and different datasets. Because our method is data-agnostic, it has the same mean and standard deviation across different datasets. Our metric has a contiguously smaller variation in the rankings of neural architectures. We believe one of the reasons is that the NASWOT method, or other data-dependent NAS methods, has a random selection of minibatches and the random initialization of model parameters. Our method, on the other hand, is not subjected to the random selection of minibatches by using fixed inputs. When we compute ranking correlations between the performance of neural architectures and graph measures, we find there is negligible difference among 8 random initialization. So we believe that our method is not significantly subjected to the different random initialization. 

\subsection{Evaluation of Different Graph Measures on NAS Benchmarks}
\label{append:graph_measures}

In addition to average degree, we also examine other graph measures in the training-free NAS. Those graph measures even exhibit a better performance than the average degree on some benchmarks.

\textbf{Graph measures.} After converting neural architectures to graphs $G(V, E)$ ($|V| = n$ and $|E| = m$) using \nasgraph{}, we compute four graph measures as new metrics to rank neural architectures in NAS benchmarks, namely, average degree, density, resilience parameter, and wedge count. \textcircled{1} The \textbf{average degree} $\bar{k}$ calculates the average number of edges for one graph node. To compute the average degree for a DAG, we ignore the direction of the graph edges. We have $\bar{k} = \frac{1}{n}\sum_{i \in V}k_{i}$, where $k_{i}$ is the degree of node $i$. \textcircled{2}  The \textbf{density} $d_{G}$ measures the ratio of the total number of edges to the maximum number of possible edges, $d_{G} = \frac{m}{n(n-1)}$. \textcircled{3}  The \textbf{resilience parameter} $\beta_{\rm{eff}}$ of a DAG \cite{gao2016universal} is defined by  $\beta_{\rm{eff}} = \frac{\mathbf{1}^T \mathbf{A} \mathbf{s}^{\rm{in}}}{\mathbf{1}^T \mathbf{A} \mathbf{1}} = \frac{\langle s^{\rm{out}}s^{\rm{in}} \rangle}{\langle s \rangle}$, where $\mathbf{1} = (1, \ldots, 1)^T$ is the all-ones vector, $\mathbf{s}^{\rm{in}} = (s^{\rm{in}}_{1}, \ldots, s^{\rm{in}}_{n})$ is the vector of incoming degrees, and $\mathbf{A}$ is the adjacency matrix of the graph. \textcircled{4} The \textbf{wedge count} $\mathcal{W}_G$ counts the number of wedges \cite{gupta2014decompositions}, and a wedge is defined as a two-hop path in an undirected graph. It is related to the triangle density of an undirected graph. To compute the wedge count of a DAG, we ignore the edge direction and use $\mathcal{W}_G = \sum_{i \in V} \frac{1}{2}k_{i}(k_{i} - 1)$. Table \ref{table:graph_measure_def} summarizes these graph measures and their computation complexity.

\begin{table*}[hbt]
\vspace{-2mm}
  \centering
  \caption{Definition and computation complexity of  the four graph measures used in \nasgraph{}.}\label{table:graph_measure_def}
  \resizebox{\textwidth}{!} {%
  \begin{tabular}{ccccc}
    \hline    
     & Average degree & Density & Resilience parameter \cite{gao2016universal} & Wedge count \cite{gupta2014decompositions} \\
    \hline
    Definition & \(\displaystyle \bar{k} = \frac{1}{n}\sum_{i \in V}k_{i} \) & \(\displaystyle d_{G} = \frac{m}{(n - 1)}\) & \(\displaystyle \beta_{\rm{eff}} = \frac{\mathbf{1}^T\mathbf{A}\mathbf{s}^{\rm{in}}}{\mathbf{1}^{T}\mathbf{A}\mathbf{1}}\) & \(\displaystyle \mathcal{W}_{G} = \sum_{i \in V} \begin{pmatrix} k_i \\ 2 \end{pmatrix} \) \\
    Time complexity & $O(m+n)$ & $O(m+n)$ & $O(n^2 + m)$ & $O(m + n)$\\
    \hline
  \end{tabular}
  }
\end{table*}

\begin{figure}[hbt]
  \centering
  \includegraphics[width=\linewidth]{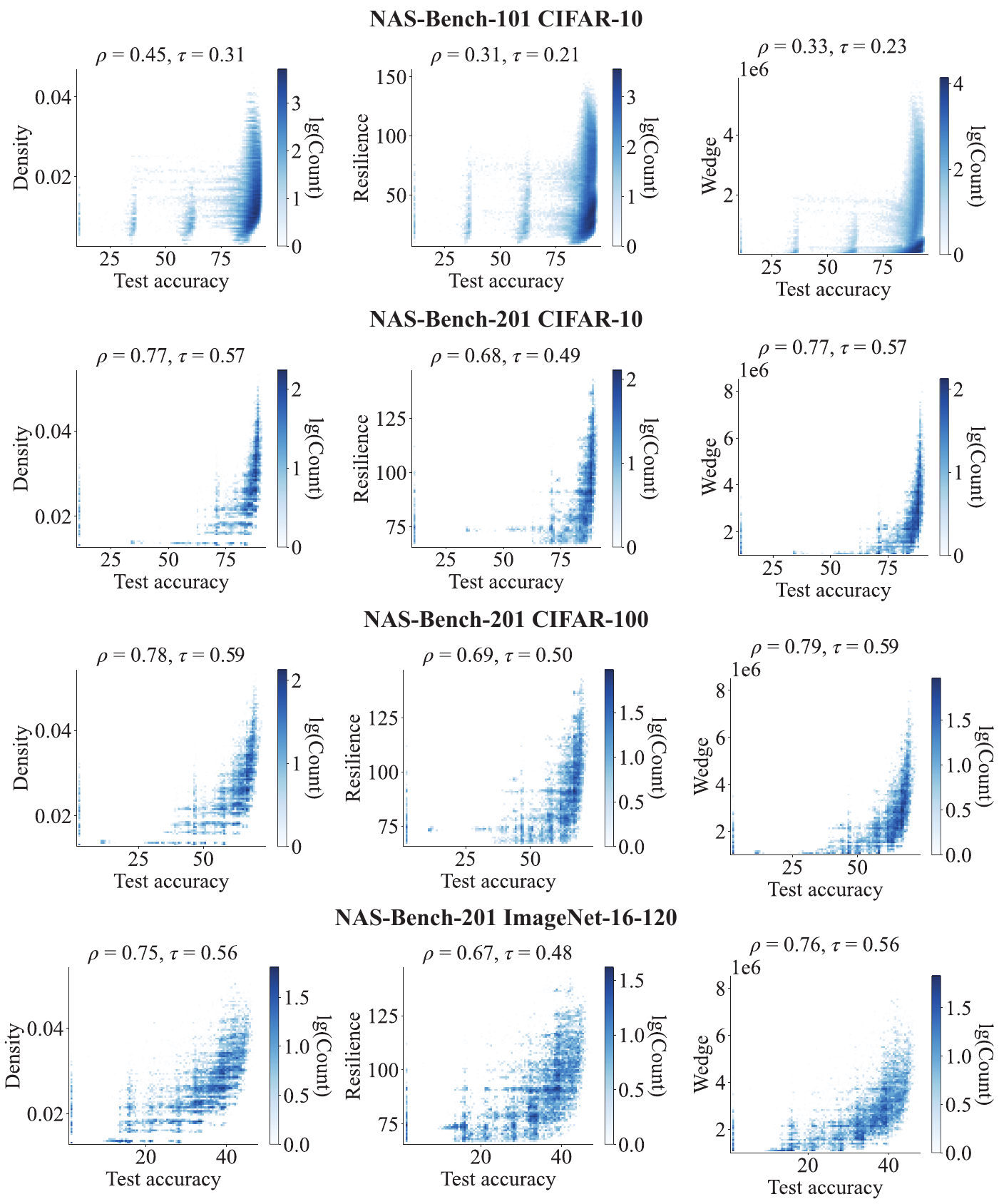}
  \vspace{-1em}
  \caption{The ranking correlations between graph measures of the converted graphs and test accuracy of the corresponding neural architectures on NAS-Bench-201. 3 graph measures (\nasgraph{} metrics) are examined: average degree (\texttt{avg\_deg}), density (\texttt{density}), resilience parameter (\texttt{resilience}) and wedge count (\texttt{wedge}).}
  \label{fig:nascor_full}
\end{figure}

Figure \ref{fig:nasbenchmark_corr} shows the evaluation on average degree and density. The ranking correlations for all 4 graph measures (average degree $\bar{k}$, density $d_G$, resilience parameter $\beta_{\rm{eff}}$ and wedge count $\mathcal{W}_{G}$) on NAS-Bench-101 and NAS-Bench-201 are shown in Figure \ref{fig:nascor_full}. The surrogate model we use is \nasgraph{(16, 1, 3)}. On NAS-Bench-101, density is the best graph measure. On NAS-Bench-201, average degree is the best graph measure across different datasets. The ranking correlation for the same graph measure does not change significantly across three datasets on NAS-Bench-201, indicating the good generality of these graph measures. The average degree and density are correlated since they are all related to the number of graph edges within a unit (total number of graph nodes or maximum number of possible edges). On NAS-Bench-201, the difference in the ranking correlation between average degree and density is marginal. However, the difference becomes larger on NAS-Bench-101. The density essentially consider the graph $G(V, E)$ as directed graph while the average degree take it as undirected graph (since we remove directionality of the graph edge). As the NAS benchmark size increases, considering $G(V, E)$ as undirected graph might be inferior given the fact that information flow in the neural architecture is directional and acyclic (Note: we are not discussing recurrent neural networks in this paper). 

\begin{table*}[hbt]
  \centering
  \caption{Comparison between the surrogate models \nasgraph{(16, 1, 3)} and \nasgraph{(4, 5, 3)} across different benchmarks and datasets.}\label{table:single_metric_compare_surrogates}
  \vspace{1em}
  \resizebox{.9\textwidth}{!} {%
  \begin{tabular}{ccccccccccccc}
    \hline
    \multirow{2}*{Methods} & \multirow{2}*{Metric} & \multicolumn{2}{c}{NAS-Bench-101} & & \multicolumn{8}{c}{NAS-Bench-201} \\
    \cline{3-4}\cline{6-13}
     & & \multicolumn{2}{c}{CIFAR-10} & & \multicolumn{2}{c}{CIFAR-10} & & \multicolumn{2}{c}{CIFAR-100} & & \multicolumn{2}{c}{ImageNet-16-120} \\
    \cline{3-4}\cline{6-7}\cline{9-10}\cline{12-13}
     & & $\rho$ & $\tau$ & & $\rho$ & $\tau$ & & $\rho$ & $\tau$ & & $\rho$ & $\tau$ \\
     \hline
     \multirow{4}*{\nasgraph{(4, 5, 3)}$^{\dagger}$} & \texttt{density} & 0.50 & 0.35 & & 0.75 & 0.56 & & 0.77 & 0.57 & & 0.75 & 0.55 \\
     & \texttt{avg\_deg} & 0.38 & 0.27 & & 0.76 & 0.57 & & 0.78 & 0.58 & & 0.76 & 0.56 \\
     & \texttt{resilience} & 0.34 & 0.23 & & 0.74 & 0.54 & & 0.75 & 0.55 & & 0.73 & 0.53 \\
     & \texttt{wedge} & 0.34 & 0.24 & & 0.76 & 0.57 & & 0.78 & 0.58 & & 0.76 & 0.56 \\
    \hline
    \multirow{4}*{\nasgraph{(16, 1, 3)}} & \texttt{density} & 0.45 & 0.31 & & 0.77 & 0.57 & & 0.78 & 0.59 & & 0.75 & 0.56 \\
     & \texttt{avg\_deg} & 0.38 & 0.26 & & 0.78 & 0.58 & & 0.80 & 0.60 & & 0.77 & 0.57 \\
     & \texttt{resilience} & 0.31 & 0.21 & & 0.68 & 0.49 & & 0.69 & 0.50 & & 0.67 & 0.48 \\
     & \texttt{wedge} & 0.33 & 0.23 & & 0.77 & 0.57 & & 0.79 & 0.59 & & 0.76 & 0.56 \\
     \hline
     \multicolumn{2}{c}{Optimal single metric} & 0.50 & 0.35 & & 0.78 & 0.58 & & 0.80 & 0.60 & & 0.77 & 0.57 \\
    \hline
  \end{tabular}%
   }
   
  \begin{tablenotes}
    \footnotesize
    \item $\dagger$ Number of cells within one module is 3 on NAS-Bench-101 while 5 on NAS-Bench-201. Because we only reduce the number channels, the surrogate model for NAS-Bench-101 is \nasgraph{(4, 3, 3)} instead of \nasgraph{(4, 5, 3)} on NAS-Bench-201.
  \end{tablenotes}
\end{table*}

\begin{table*}[hbt]
  \centering
  \caption{Ranking correlations $\rho$ between the validation accuracies and \nasgraph{} metrics using different surrogate models. Because the structure of modules and cells is encoded in the arch string such as \texttt{64-41414-1\_02\_333} (we refer the reader to \protect\cite{duan2021transnas} for details on the arch string), we only change the number of channels.}\label{table:single_metric_transbench101_surrogates}
  \vspace{1em}
  \resizebox{0.76\textwidth}{!}{%
  \begin{tabular}{cccccccccccc}
    \hline
    \multirow{3}*{Metric} & \multicolumn{11}{c}{Micro TransNAS-Bench-101} \\
    \cline{2-12}
     & \multicolumn{2}{c}{\texttt{class\_object}} & & \multicolumn{2}{c}{\texttt{class\_scene}} & & \multicolumn{2}{c}{\texttt{room\_layout}} & & \multicolumn{2}{c}{\texttt{segment\_semantic}}\\
     \cline{2-3}\cline{5-6}\cline{8-9}
     & $\rho$ & $\tau$ & & $\rho$ & $\tau$ & & $\rho$ & $\tau$ & & $\rho$ & $\tau$\\
    \hline
    \multicolumn{12}{c}{4 Channels} \\
    \texttt{avg\_deg} & 0.56 & 0.39 & & 0.71 & 0.51 & & 0.38 & 0.25 & & 0.67 & 0.48 \\
    \texttt{density} & 0.54 & 0.37 & & 0.68 & 0.49 & & 0.37 & 0.24 & & 0.60 & 0.43 \\
    \texttt{resilience} & 0.62 & 0.44 & & 0.75 & 0.55 & & 0.47 & 0.31 & & 0.33 & 0.23 \\
    \texttt{wedge} & 0.60 & 0.42 & & 0.74 & 0.54 & & 0.43 & 0.28 & & 0.68 & 0.49 \\
    \hline
    \multicolumn{12}{c}{16 Channels} \\
    \texttt{avg\_deg} & 0.55 & 0.38 & & 0.70 & 0.50 & & 0.37 & 0.24 & & 0.66 & 0.47 \\
    \texttt{density} & 0.53 & 0.36 & & 0.68 & 0.48 & & 0.35 & 0.22 & & 0.55 & 0.39\\
    \texttt{resilience} & 0.62 & 0.43 & & 0.74 & 0.54 & & 0.47 & 0.31 & & 0.34 & 0.24 \\
    \texttt{wedge} & 0.59 & 0.41 & & 0.74 & 0.54 & & 0.42 & 0.27 & & 0.68 & 0.49 \\
    \hline
  \end{tabular}
  }
\end{table*}

As indicated in Figure \ref{fig:ablation_3d}, decreasing number of cells or number of channels does not cause a huge effect on the rankings of graph measures. In addition to reduce the number of cells (the surrogate model \nasgraph{(16, 1, 3)}), we also examine the effect of reducing the number of channels (the surrogate model \nasgraph{(4, 5, 3)}). The comparison of the ranking correlation using different surrogate models is shown in the Table \ref{table:single_metric_compare_surrogates}. These two surrogate models, as expected, have a similar ranking correlation across NAS-Bench-101 and NAS-Bench-201.

In addition to these two benchmarks, we also examine the performance of different surrogate model on TransNAS-Bench-101. Because the number of cells and modules are fixed in the arch string (the way to represent neural architecture on the benchmark), we only change the number of channels. Table \ref{table:single_metric_transbench101_surrogates} shows the evaluation of two surrogate models. There are only marginal difference between two graph measures across three tasks.

Overall, as indicated in the ablation study, decreasing number of channels and decreasing number of cells do not have a significant change in the rankings of graph measures. Based on graph theory, decreasing number of cells is preferred.

\subsection{Combination of Training-free NAS Metrics}
\label{append:metric_comb}

\begin{figure}[hbt]
  \centering
  \includegraphics[width=0.96\linewidth]{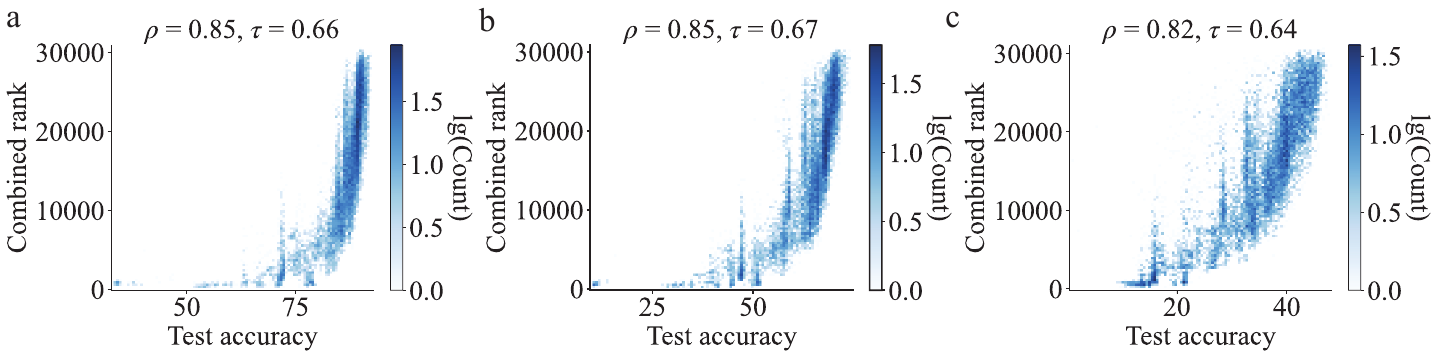}
  \caption{Combined rank (rank(\texttt{avg\_deg}) + rank(\texttt{jacob\_cov})) of neural architectures of NAS-Bench-201 vs test accuracy. (a) CIFAR-10. (b) CIFAR-100. (c) ImageNet-16-120.}
  \label{fig:metric_combine}
\end{figure}

We combine graph measure with training-free NAS metrics (rank(\texttt{avg\_deg}) + rank(\texttt{jacob\_cov})) to boost the ranking correlation between metrics and performance of neural architectures. Figure \ref{fig:metric_combine} shows the relationship between combined ranks and test accuracies across CIFAR-10, CIFAR-100 and ImageNet-16-120 datasets. We note that \texttt{avg\_deg} is data-agnostic while \texttt{jacob\_cov} is data-dependent.

\subsection{Random Search in NAS}
\label{append:random_search}

Algorithm \ref{alg:random_search_alg} shows the process of random search using a single metric. A total number of $N$ neural architectures are randomly sampled from the same NAS benchmark such as NAS-Bench-101. Metrics are computed as the score by a single forward/backward propagation of neural architectures with randomly initialized parameters. In the \nasgraph{} framework, we convert neural architectures to graphs and then compute graph measures such as average degree to rank neural architectures. The performance, e.g. test accuracy, of the neural architecture with the highest graph measure is extracted as the performance of the metric. The highest performance of the selected $N$ architectures are used as GT.

\begin{algorithm}[htb]
\caption{Random Search Algorithm Using Single Metric}\label{alg:random_search_alg}
\begin{algorithmic}[1]
  \STATE net\_generator = RandomGenerator()
  \STATE score\_highest, net\_best = None, 0
  \FOR {$i = 1 : N$}
    \STATE net = net\_generator.pick\_net()
    \STATE score = ComputeMetric(net)
    \IF {score $>$ score\_highest}
      \STATE score\_highest = score
      \STATE net\_best = net
    \ENDIF
  \ENDFOR
  \STATE acc\_best = ExtractAccFromBenchmark(net\_best)
\end{algorithmic}
\end{algorithm}

\subsection{Cell Structure}
\label{append:cell_struct}
\begin{figure}[hbt]
  \includegraphics[width=\linewidth]{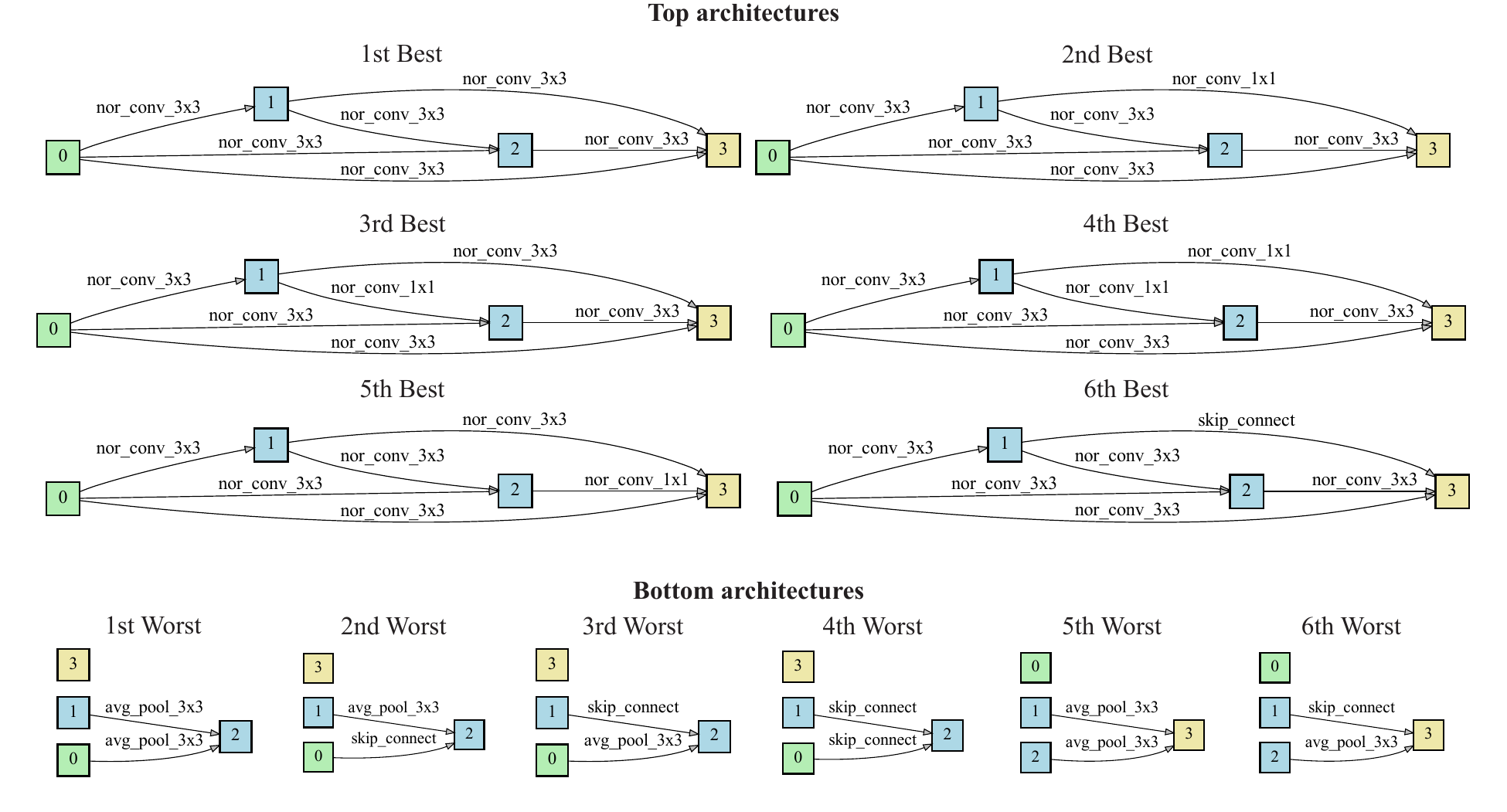}
  \caption{The top 6 best and top 6 worst cell structures ranked by average degree of \nasgraph{} on NASBench-201. Architectures are ranked by the graph measure \texttt{avg\_deg}.}
  \label{fig:nas201archview}
\end{figure}

We visualize the best and the worst cell structures ranked by \texttt{avg\_deg} on NAS-Bench-201 as shown in Figure \ref{fig:nas201archview}. Edge with \texttt{none} operation is not shown for better visualization. The best cell structure found by our metric is same as \texttt{synflow}. The worst cell structures share a same feature: there is an isolated node of cell structure. The isolated node in NAS-Bench-201 means extracted features from preceding neural layers are disregarded, and the effective depth of neural architecture becomes shallower. Therefore, it is expected those architectures have a poor performance.

\subsection{Special NASGraphs}
\label{append:special_graphs}

\begin{figure}[hbt]
  \centering
  \includegraphics[width=0.96\linewidth]{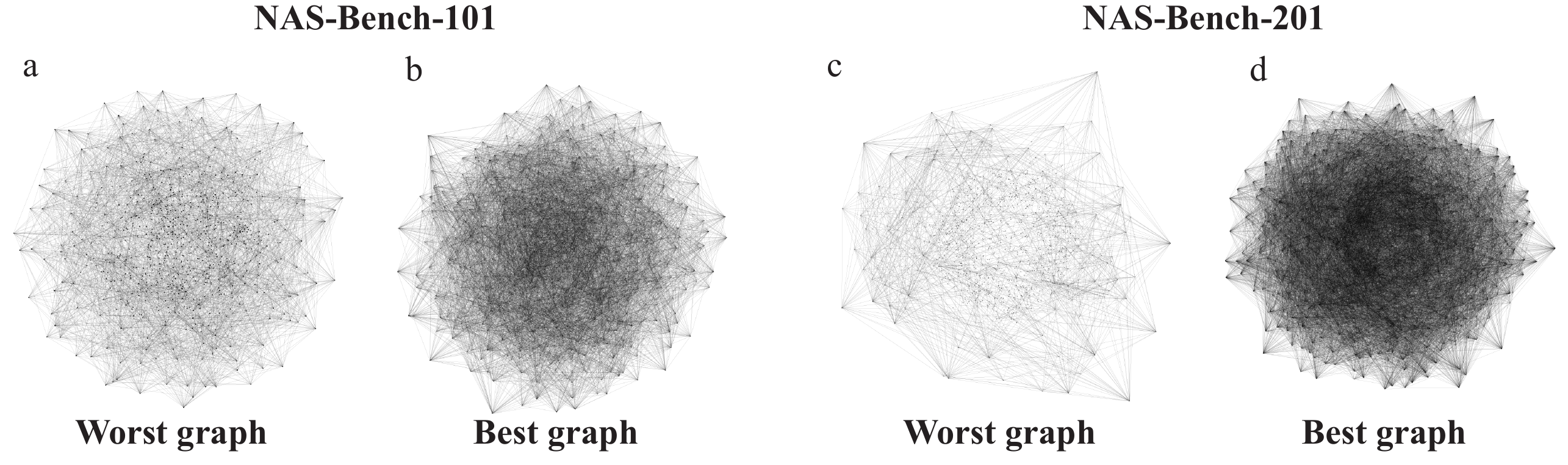}
  \caption{The converted graphs corresponding to the best architecture and the worst architecture on NAS-Bench-101 and NAS-Bench-201. Architectures are ranked by the test accuracy.}
  \label{fig:special_graphs}
\end{figure}

Figure \ref{fig:special_graphs} shows the visualization of the converted graphs using the surrogate model \nasgraph{(4, 3, 5)} corresponding to the best architecture and the worst architecture (ranked by the test accuracy) in NAS-Bench-101 and NAS-Bench-201. Instead of ranking architectures by the graph measures (as the ranking metric in Figure \ref{fig:nasbenchmark_corr}), we rank architectures by the test accuracy. As indicated by graph measures, the best graph is much denser compared to the worst graph.

\subsection{Supplementary Results on NAS Benchmarks}

In addition to NAS-Bench-201, TransNAS-Bench-101 and NDS, we also compare our method with NAS-Bench-101, the result is shown in the Table \ref{table:nasbench101_comp}. Our proposed method has a performance close to the SOTA method.

\begin{table}[H]
  \centering
  \caption{Comparison of the ranking correlation between \nasgraph{} and training-free NAS methods on the NAS-Bench-101 benchmark.}\label{table:nasbench101_comp}
  \vspace{1em}
  \resizebox{.42\textwidth}{!} {%
  \begin{tabular}{ccccccccccccc}
    \hline
    \multirow{2}*{Method} & \multirow{2}*{Metric} & \multicolumn{2}{c}{NAS-Bench-101}\\
    \cline{3-4}
     & & \multicolumn{2}{c}{CIFAR-10} \\
    \cline{3-4}
     & & $\rho$ & $\tau$ \\
     \hline
    NASWOT & relu\_logdet & - & \textbf{0.31}\\
    ZiCo & \texttt{zico} & \textbf{0.45} & \textbf{0.31} \\
    \hline
    \multirow{6}*{Zero-Cost NAS} & \texttt{grad\_norm} & 0.20 & - \\
     & \texttt{snip} & 0.16 & - \\
     & \texttt{grasp} & 0.45 & - \\
     & \texttt{fisher} & 0.26 & - \\
     & \texttt{synflow} & 0.37 & - \\
     & \texttt{jacob\_cov} & 0.38 & - \\
    \hline
    Ours & \texttt{avg\_deg} & 0.38 & 0.26 \\
    \hline
  \end{tabular}%
  }
\end{table}





\end{document}